\documentclass{article}
\usepackage{iclr2024,times}

\usepackage{hyperref,url}
\usepackage{subfig,tabularx,graphicx}
\usepackage{amssymb,amsmath,amsfonts}
\usepackage{comment}
\usepackage{mathtools}
\usepackage{booktabs}
\usepackage{geometry,multicol,multirow}
\usepackage{algorithm,algpseudocode}
\usepackage{scalerel}
\usepackage{wrapfig}
\hypersetup{colorlinks,linkcolor=red,citecolor=teal,urlcolor=blue}

\usepackage{pifont}
\newcommand{\cmark}{\textcolor{green!60!black}{\ding{51}}}
\newcommand{\xmark}{\textcolor{red!60!black}{\ding{55}}}

\def\x{\mathbf{x}}
\def\u{\mathbf{u}}
\def\v{\mathbf{v}}
\def\y{\mathbf{y}}

\def\t{\mathbf{t}}

\def\grayxt{\textcolor{gray}{(\mathbf{x},t)}}
\def\grayt{\textcolor{gray}{(t)}}
\def\graytau{\textcolor{gray}{(\tau)}}
\def\graytz{\textcolor{gray}{(t_0)}}
\def\grayti{\textcolor{gray}{(t_i)}}
\def\bmu{\boldsymbol{\mu}}
\def\bs{\boldsymbol{\sigma}}
\def\0{\mathbf{0}}
\def\dv{\dot{\mathbf{v}}}
\def\R{\mathbb{R}}
\def\E{\mathbb{E}}
\def\N{\mathcal{N}}
\def\L{\mathcal{L}}
\def\D{\mathcal{D}}

\DeclareMathOperator*{\argmin}{arg\,min}

\title{ClimODE: Climate and Weather Forecasting \\ with Physics-informed Neural ODEs}

\author{Yogesh Verma, Markus Heinonen \\
Department of Computer Science\\
Aalto University, Finland \\
\texttt{\{yogesh.verma,markus.o.heinonen\}@aalto.fi}
\And Vikas Garg\\
YaiYai Ltd and Aalto University\\
\texttt{vgarg@csail.mit.edu}
}

\iclrfinalcopy 

\begin{document}

\maketitle

\begin{abstract}
Climate and weather prediction traditionally relies on complex numerical simulations of atmospheric physics. Deep learning approaches, such as transformers, have recently challenged the simulation paradigm with complex network forecasts. However, they often act as data-driven black-box models that neglect the underlying physics and lack uncertainty quantification. We address these limitations with ClimODE, a  spatiotemporal continuous-time process that implements a key principle of \emph{advection} from statistical mechanics, namely, weather changes due to a spatial movement of quantities over time. ClimODE models precise weather evolution with value-conserving dynamics, learning global weather transport as a neural flow, which also enables estimating the uncertainty in predictions. Our approach outperforms existing data-driven methods in global and regional forecasting with an order of magnitude smaller parameterization, establishing a new state of the art.

\end{abstract}

\section{Introduction}

State-of-the-art climate and weather prediction relies on high-precision numerical simulation of complex atmospheric physics \citep{phillips1956general,satoh2004atmospheric,lynch2008origins}. While accurate to medium timescales, they are computationally intensive and largely proprietary \citep{gfs,ifs}.

%The state-of-the-art for climate prediction is heavy numerical simulation methods based on careful understanding, derivation and simulation of atmospheric physics \citep{phillips1956general,satoh2004atmospheric,lynch2008origins}. While very accurate at near to medium timescales, these models are highly complex, computationally expensive, and often proprietary (CITE).

There is a long history of `free-form' neural networks challenging the mechanistic simulation paradigm \citep{kuligowski1998localized,baboo2010efficient}, and recently deep learning has demonstrated significant successes \citep{nguyen2023climax}. These methods range from one-shot GANs \citep{ravuri2021skilful} to autoregressive transformers \citep{pathak2022fourcastnet,nguyen2023climax,bi2022panguweather} and multi-scale GNNs \citep{lam2022graphcast}. \citet{zhang2023skilful} combines autoregression with physics-inspired transport flow. %Several of these models are proprietary as well.

%A seminal work of \citet{ravuri2021skilful} performs one-shot prediction with a GAN. Series of autoregressive models predict next timepoint recurrently with transformers either in ambient `pixel' space \citep{nguyen2023climax} or in Fourier domain \citep{pathak2022fourcastnet}, while GraphCast include multi-scale modelling with GNNs \citep{lam2022graphcast}. NowcastNet augments the autoregression with incompressible Navier-Stokes term \citep{zhang2023skilful}. The above methods are based on discrete-time predictive `jumps' forward in time, which incurs numerical approximations. The climate is also an open system that is compressible.

\begin{figure}[t]
    \centering
    \includegraphics[width=\textwidth]{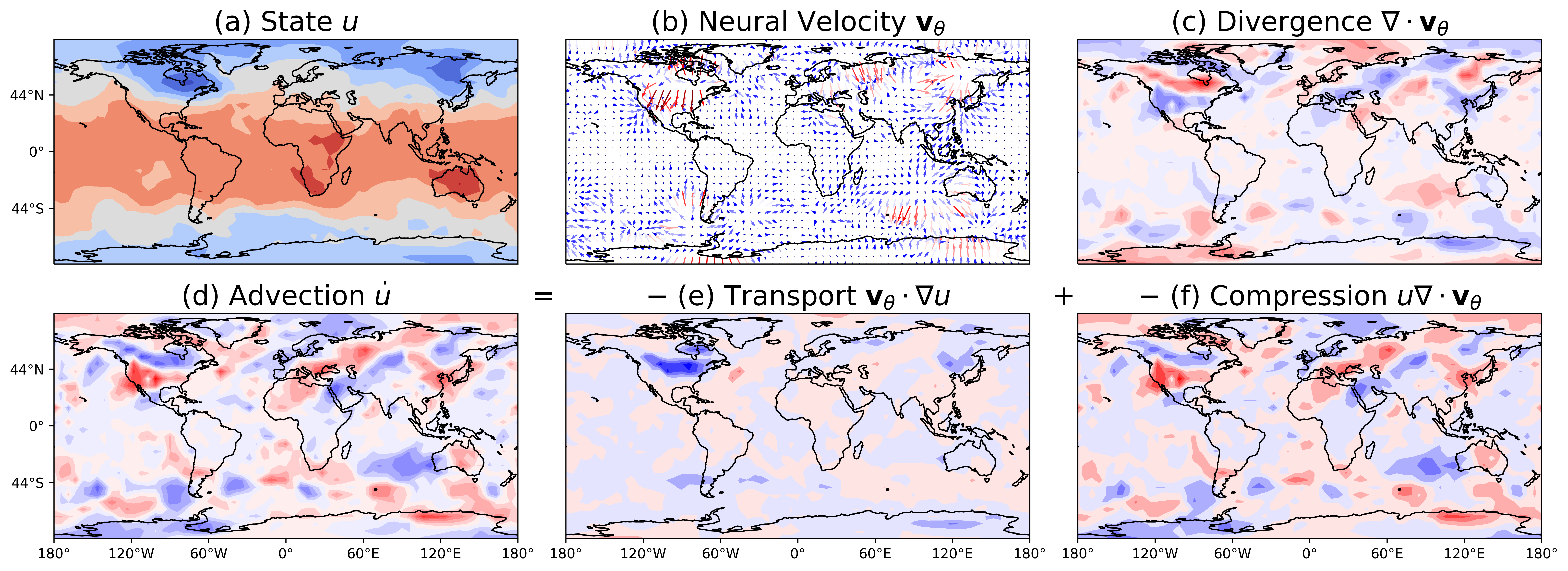}
    \caption{\textbf{Weather as a quantity-preserving advection system.} A quantity (eg. temperature) \textbf{(a)} is moved by a neural flow velocity \textbf{(b)}, whose divergence is the flow's compressibility \textbf{(c)}. The flow translates into state change by advection \textbf{(d)}, which combine quantity's transport \textbf{(e)} and compression \textbf{(f)}.}
    \label{fig:system}
\end{figure}

In statistical mechanics, weather can be described as a \emph{flux}, a spatial movement of quantities over time, governed by the partial differential \emph{continuity equation} \citep{broome2014pde}
\begin{align} \label{eq:ce}
    \underbrace{\frac{du}{dt}}_{\text{time evolution } \dot{u}} + \:\: \underbrace{ \overbrace{\v \cdot \nabla u}^{\text{transport}} \: + \overbrace{u \nabla \cdot \v}^{\text{compression}}}_{\text{advection}} = \underbrace{s}_{\text{sources}},
\end{align}
where $u(\x,t)$ is a quantity (e.g. temperature) evolving over space $\x \in \Omega$ and time $t \in \R$ driven by a flow's velocity $\v(\x,t) \in \Omega$ and sources $s(\x,t)$ (see Figure \ref{fig:system}). The advection moves and redistributes existing weather `mass' spatially, while sources add or remove quantities. Crucially, the dynamics need to be continuous-time, and modeling them with autoregressive `jumps' violates the conservation of mass and incurs approximation errors. %\color{red} V: we need to be consistent with our nomenclature: e.g., sometimes we use transport including negative sign and sometimes without it; sometimes we use term compression, other times, divergence for the same thing.  Let's also take care of "Concentration" in Figure 1. This might help: \url{https://climate.ucdavis.edu/ATM121/AtmosphericDynamics-Chapter01-Part03-Continuity.pdf}  \color{black} \yv{I am searching literature, will change it} 

We introduce a climate model that implements a continuous-time, second-order neural continuity equation with simple yet powerful inductive biases that ensure -- by definition -- value-conserving dynamics with more stable long-horizon forecasts. We show a computationally practical method to solve the continuity equation over entire Earth as a system of neural ODEs. We learn the flow $\v$ as a neural network with only a few million parameters that uses both global attention and local convolutions. Furthermore, we address source variations via a probabilistic emission model that quantifies prediction uncertainties. Empirical evidence underscores ClimODE's ability to attain state-of-the-art global and regional weather forecasts.

\subsection{Contributions}
We propose to learn a continuous-time PDE model, grounded on physics, for climate and weather modeling and uncertainty quantification. In particular,
\begin{itemize}
    \item we propose ClimODE, a continuous-time neural advection PDE climate and weather model, and derive its ODE system tailored to numerical weather prediction.
    \item we introduce a flow velocity network that integrates local convolutions, long-range attention in the ambient space, and a Gaussian emission network for predicting uncertainties and source variations.
    \item empirically, ClimODE achieves state-of-the-art global and regional forecasting performance.
    \item Our physics-inspired model enables efficient training from scratch on a single GPU and comes with an open-source PyTorch implementation on GitHub.\footnote{\url{https://github.com/Aalto-QuML/ClimODE}}
    %\item Due to the physics-inspired model construction, the neural networks are compact and (re-)trainable from scratch with a single ordinary GPU (e.g. V100) in hours. We publish an open Pytorch implementation at Github\footnote{Link anonymised during review period.}
\end{itemize}

\section{Related works}

\begin{table}[ht]
    \caption{ \textbf{Overview of current deep learning methods for weather forecasting.} }
    %\color{red} V: How do we get these parameters? Not all the methods are using the same data, and the number of parameters they use might depend on data? Also, do the number of reported parameters for ClimaX here include parameters from pretraining (since we don't use any pretraining in our experiments)? \color{black} \yv{These param are taken from their official papers. ClimaX use the same param first pretrain it and then use same to finetune.}}
    \label{tab:comp_table}
    \vskip 0.05in
    \begin{center}
            \resizebox{\textwidth}{!}{\begin{tabular}{lcccccr}
                \toprule
                Method & Value-preserving & Explicit Periodicity/Seasonality & Uncertainty & Continuous-time & Parameters (M) & \\ \midrule
                FourCastNet     & \xmark & \xmark & \xmark& \xmark & N/A &\citet{pathak2022fourcastnet}     \\
                GraphCast  & \xmark & \xmark & \xmark & \xmark & $37$ &\citet{lam2022graphcast} \\ % 36.7           
                Pangu-Weather     & \xmark & \xmark & \xmark& \xmark & $256$ &\citet{bi2022panguweather}     \\
                ClimaX    & \xmark & \xmark &\xmark & \xmark & $107$ & \citet{nguyen2023climax}     \\  
                NowcastNet    & \cmark & \xmark & \xmark &\xmark & N/A &\citet{zhang2023skilful}      \\
                \midrule
                ClimODE    & \cmark & \cmark & \cmark &\cmark & $2.8$ &this work    \\
                \bottomrule
            \end{tabular}}
    \end{center}
    \vskip -0.1in
\end{table}

\paragraph{Numerical climate and weather models.} 
Current models %rely heavily on computational methods to predict Earth's weather systems based on a common set of primitive equations. These methods 
encompass numerical weather prediction (NWP) for short-term weather forecasts and climate models for long-term climate predictions. The cutting-edge approach in climate modeling involves earth system models (ESM) \citep{hurrell2013community}, which integrate simulations of physics of the atmosphere, cryosphere, land, and ocean processes. %ESMs, dating back to solving Navier-Stokes equations for fluid circulation~\citep{satoh2004atmospheric,lynch2008origins} on a rotating sphere, serve as vital tools for studying climate variations, particularly in response to factors like greenhouse gases. 
While successful, they exhibit sensitivity to initial conditions, structural discrepancies across models~\citep{balaji2022general}, regional variability, and high computational demands.

%Current research at the intersection of weather and climate science and ML has largely focused on designing separate models for every task of interest despite potential availability of fairly diverse large scale data with shared underlying physics and geology across these tasks. 

\paragraph{Deep learning for forecasting.} Deep learning has emerged as a compelling alternative to NWPs, focusing on global forecasting tasks.  \citet{rasp2020weatherbench} employed pre-training techniques using ResNet~\citep{he2016deep} for effective medium-range weather prediction, \citet{weyn2021sub} harnessed a large ensemble of deep-learning models for sub-seasonal forecasts, whereas \citet{ravuri2021skilful} used deep generative models of radar for precipitation nowcasting and GraphCast~\citep{lam2022graphcast,keisler2022forecasting} utilized a graph neural network-based approach for weather forecasting. Additionally, recent state-of-the-art neural forecasting models of ClimaX~\citep{nguyen2023climax}, FourCastNet~\citep{pathak2022fourcastnet}, and Pangu-Weather~\citep{bi2022panguweather} are predominantly built upon data-driven backbones such as Vision Transformer (ViT)~\citep{dosovitskiy2021image}, UNet~\citep{ronneberger2015u}, and autoencoders. However, these models overlook the fundamental physical dynamics and do not offer uncertainty estimates for their predictions.

%Finally, recent state-of-the-art neural forecasting models such as ClimaX~\citep{nguyen2023climax}, FourCastNet~\citep{pathak2022fourcastnet}, Pangu-Weather~\citep{bi2022panguweather} are largely based on data-driven backbone like Vision Transformer (ViT) \citep{dosovitskiy2021image}, UNet \citep{ronneberger2015u}, and autoencoders.  %Unlike these works, we formulate a physics-driven backbone in our model to accommodate the underlying physical dynamics as neural physics ODEs, providing an uncertainty estimate tailored to numerical weather prediction.

\paragraph{Neural ODEs.} Neural ODEs propose learning the time derivatives as neural networks \citep{chen2019neural,massaroli2020dissecting}, with multiple extensions to adding physics-based constraints \citep{greydanus2019hamiltonian,cranmer2020lagrangian,brandstetter2022clifford,choi2023climate}. The physics-inspired networks (PINNs) embed mechanistic understanding in neural ODEs \citep{raissi2019physics,cuomo2022scientific}, while multiple lines of works attempt to uncover interpretable differential forms \citep{brunton2016discovering,Fronk_2023}.  Neural PDEs warrant solving the system through spatial discretization \citep{poli2019graph,iakovlev2020learning} or functional representation \citep{li2020fourier}. Machine learning has also been used to enhance fluid dynamics models \citep{li2020fourier,lu2021learning,kochkov2021machine}. The above methods are predominantly applied to only small, non-climate systems. % commonly applied to only small systems and, at the current time, have not been applied to weather. 

%To our knowledge, there are no neural ODE models that opt explicit fluid dynamics structures, nor neural ODEs for climate data. (TODO)

%\paragraph{Statistical mechanics.} The climate can be seen as a large particle system, where the particles move under a flow \citep{broome2014pde}. The evolution of the particle density follows Fokker-Planck equations, or Navier-Stokes equations in fluid dynamics \citep{kreyszig2008advanced}. These methods primarily consider diffusion and advection as a partial differential equation 
%\begin{align}
%    \frac{du}{dt} = \Delta u + \nabla \cdot (u\v) + s, 
%\end{align}
%where $u$ is a variable of interest carried by flow velocity $\v$, dissipated by the Laplacian $\Delta = \nabla^2$, and added or removed by $s$.

\section{Neural transport model}

\paragraph{Notation.}
Throughout the paper $\nabla = \nabla_\x$ denotes spatial gradients, $\dot{u} = \frac{du}{dt}$ time derivatives, $\cdot$ inner product, and $\nabla \cdot \v = \operatorname{tr} (\nabla \v)$ divergence. We color equations purely for cosmetic clarity.

%\subsection{Primer to fluid dynamics}

%Fluid dynamics is a branch of statistical physics that studies the mechanics of potentials under flow motion. The dominant formalism of Navier-Stokes represents fluids by conservation of mass, energy, and momentum. The continuity equation is at the core of fluid dynamics, often explained as the temperature change in a pool of water caused by a current. (etcetc)

%In this work, we do not explicitly model dissipation or stochasticity, which are estimated to be minor effects at global scale (CITE). We also ignore forcings (such as pollution, volcanoes or astronomic effects) as standard benchmarks lack support for them.

\begin{figure}[t]
    \centering
    \includegraphics[width=0.71\textwidth]{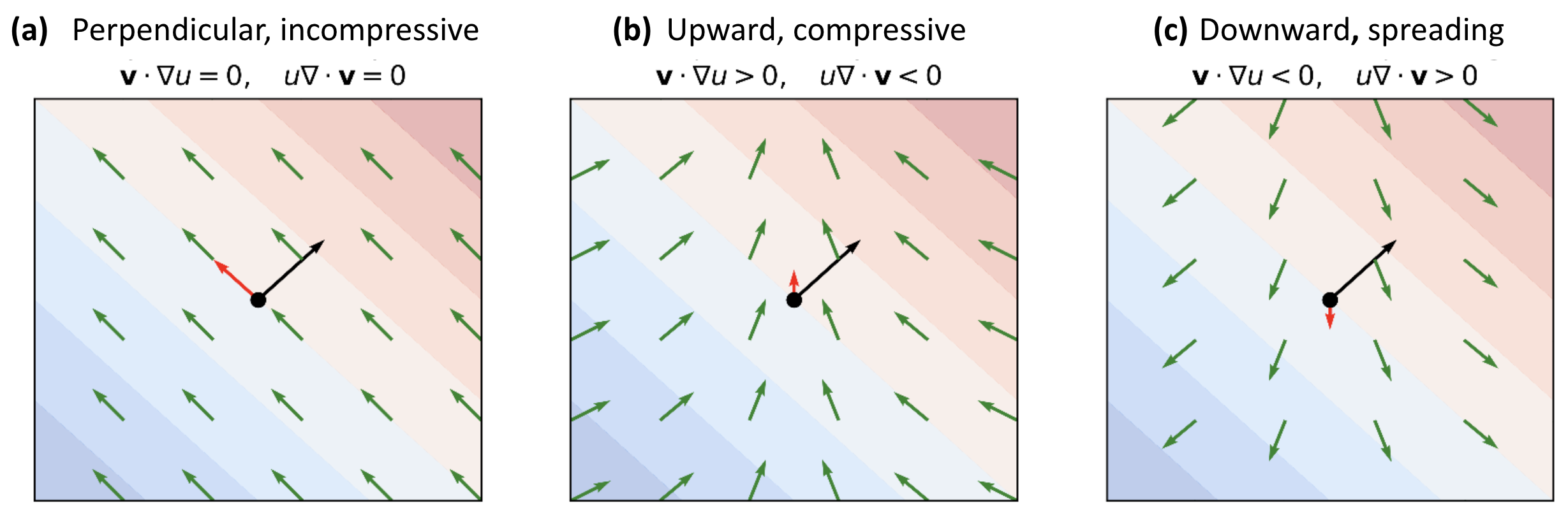}
    \caption{\textbf{Conceptual illustration of  continuity equation on pointwise temperature change} $\dot{u}(\x_0,t) = - \v \cdot \nabla u - u\nabla \cdot \v$. \textbf{(a)} A perpendicular flow (\textcolor{green!60!black}{green}) to the gradient (\textcolor{blue!60!white}{blue} \textcolor{red!40!white}{to} \textcolor{red!60!white}{red}) moves in equally hot air causing no change at $\x_0$. \textbf{(b)} Cool air moves upwards, decreasing pointwise temperature, while air concentration at $\x_0$ accumulates additional temperature. \textbf{(c)} Hot air moves downwards increasing temperature at $\x_0$, while air dispersal decreases it. }
    %\color{red}  V: In real world, doesn't the hot air move upwards causing inclement (cold)  weather, while cool air moves downwards because it is more dense than hot air? It also seems the figure is trimmed from the top, resulting in issues with the text. \color{black} \yv{This is an illustration depending on the heatmap and the gradient related to it, not like in a room setting.} }
    \label{fig:continuity}
\end{figure}

\subsection{Advection equation}

We model weather as a spatiotemporal process $\u\grayxt = (u_1\grayxt, \ldots, u_K\grayxt) \in \R^K$ of $K$ quantities $u_k\grayxt \in \R$ over continuous time $t \in \R$ and latitude-longitude locations $\x = (h,w) \in \Omega = [-90^\circ, 90^\circ] \times [-180^\circ, 180^\circ] \subset \R^2$. %We consider $K=5$ quantities from the ERA5 dataset: ground temperature (\texttt{t2m}), atmospheric temperature (\texttt{t}), geopotential (\texttt{z}), and ground wind vector (\texttt{u10}, \texttt{v10}). We assume a spatial discretisation into a grid of resolution $(32,64)$ with spacing of $5.625^\circ$, while the methods apply to arbitrary resolutions. See Appendix \ref{sec:data} for details.
We assume the process follows an advection partial differential equation 
% over a $(H=32,W=64)$ grid and timepoints $(t_1,\ldots, t_N)$ with 6 hour increments $\Delta t = 0.25$.
%Specifically, we propose a 2nd order neural transport evolution model,
\begin{align} \label{eq:pde}
    \dot{u}_k\grayxt %&= - \nabla \cdot \big( u_k\grayxt \v_k\grayxt \big) \\
    &= - \underbrace{\v_k\grayxt \cdot \nabla u_k\grayxt}_{\text{transport}} - \underbrace{u_k\grayxt \nabla \cdot \v_k\grayxt}_{\text{compression}}~,
\end{align}
where quantity change $\dot{u}_k\grayxt$ is caused by the flow, whose velocity $\v_k\grayxt \in \Omega$ transports and concentrates air mass (see Figure \ref{fig:continuity}). The equation \eqref{eq:pde} describes a \emph{closed} system, where value $u_k$ is moved around but never lost or added. While a realistic assumption on average, we will introduce an emission source model in Section \ref{sec:uncertain}. % To account for local transfers, such as night-time heat radiation to space, we will introduce a For instance, the sun's heating effect is balanced by heat radiation to space on a global scale, while heat transfer between the atmosphere and earth is balanced over time~\citep{sellers1969global}. 
The closed system assumption forces the simulated trajectories $u_k\grayxt$ to \emph{value-preserving} manifold 
%This preservation assumption is approximately true at macroscopic level (CITE).
\begin{align}
    \int u_k(\x,t)d\x &= \mathrm{const}, \qquad \forall t,k.
\end{align}
This is a strong inductive bias that prevents long-horizon forecast collapses (see Appendix \ref{subsec:mass_consv} for details.)

\subsection{Flow velocity}

Next, we need a way to model the flow velocity $\v\grayxt$ (See Figure \ref{fig:system}b). Earlier works have remarked that second-order bias improves the performance of neural ODEs significantly \citep{yildiz2019ode2vae,gruver2022deconstructing}. Similarly, we propose a second-order flow by parameterizing the change of velocity with a neural network $f_\theta$,
\begin{align} \label{eq:velocity}
    \dv_k\grayxt &= f_\theta\Big(\u\grayt, \nabla \u\grayt, \v\grayt, \psi\Big),
\end{align}
as a function of the current state $\u\grayt = \{ \u\grayxt : \x \in \Omega\} \in \R^{K \times H \times W}$, its gradients $\nabla \u\grayt \in \R^{2K \times H \times W}$, the current velocity $\v\grayt = \{ \v\grayxt : \x \in \Omega\} \in \R^{2K \times H \times W}$, and spatiotemporal embeddings $\psi \in \R^{C \times H \times W}$. These inputs denote global \emph{frames} (e.g., Figure \ref{fig:system}) at time $t$ discretized to a resolution $(H,W)$ with a total of $5K$ quantity channels and $C$ embedding channels.
\subsection{2nd-order PDE as a system of first-order ODEs}
%PDE solutions, represented by $u(\mathbf{x},t)$ with initial conditions $u(\mathbf{x},t_0)$, are often complex due to their dependence on both time $t$ and space $\mathbf{x}$. In contrast, ODEs only depend on time, allowing straightforward numerical forward 'unrolling' \citep{evans2022partial}. To handle this, 
We utilize the method of lines (MOL), discretizing the PDE into a grid of location-specific ODEs \citep{schiesser2012numerical,iakovlev2020learning}. Additionally, a second-order differential equation can be transformed into a pair of first-order differential equations \citep{kreyszig2008advanced,yildiz2019ode2vae}. Combining these techniques yields a system of first-order ODEs $(u_{ki}(t), \mathbf{v}_{ki}(t))$ of quantities $k$ at locations $\mathbf{x}_i$:
%PDE solutions $u(\x,t)$ from initial conditions $u(\x,t_0)$ are often unwieldy both in theory and in practise due to being dependent on both time $t$ and space $\x$. In contrast, ordinary differential equations (ODEs) depend only on time, and admit simple numerical forward `unrolling' \citep{evans2022partial}. Thus, we use method of lines (MOL) to discretise a PDE into a grid of location-specific ODEs  \citep{schiesser2012numerical,iakovlev2020learning}. Furthermore, a second-order differential equation can be equivalently expanded into a system of two first-order differential equations \citep{kreyszig2008advanced,yildiz2019ode2vae}. Applying both techniques results in a system of first-order ODEs $(u_{ki}(t), \v_{ki}(t))$ for a quantity $k$ at location $\x_i$,
%
\begin{align} \label{eq:solution}
    \begin{bmatrix} 
     \u\grayt \\ 
     \v\grayt 
    \end{bmatrix} 
    = 
    \begin{bmatrix} 
     \u\graytz \\ 
     \v\graytz 
    \end{bmatrix} 
    + \int_{t_0}^t 
    \begin{bmatrix} 
     \dot{\u}\graytau \\ 
     \dv\graytau \end{bmatrix} 
    d\tau 
    = 
    \begin{bmatrix} 
    \big\{ u_k\graytz \big\}_k \\ 
    \big\{ \v_k\graytz \big\}_k 
    \end{bmatrix} 
    + \int_{t_0}^t
    \begin{bmatrix} 
     \big\{ -\nabla \cdot (u_k\graytau \v_k\graytau) \big\}_k \\ 
     \big\{ f_\theta\big( \u\graytau, \nabla \u\graytau, \v\graytau, \psi\big)_k \big\}_k
    \end{bmatrix} 
    d\tau,
\end{align}
where $\tau \in \R$ is an integration time, and where we apply equations \eqref{eq:pde} and \eqref{eq:velocity}. Backpropagation of ODEs is compatible with standard autodiff, while also admitting tractable adjoint form \citep{lecun1988theoretical,chen2019neural,metz2021gradients}. The forward solution $\u(t)$ can be accurately approximated with numerical solvers such as Runge-Kutta \citep{runge1895numerische} with low computational cost. 

%In our experiments we assume $K=5$ quantities and spatial discretisation of earth to resolution $(H,W) = (32,64)$ resulting in a total of $3KWH = 30720$ scalar ODEs. This can seem daunting, but they all \emph{share the same differential function} $f_\theta$, that is, the time evolution at Tokyo and at New York follows the same rules. The system can then be batched into a single image %$\mathrm{stack}[\u\grayt;\v\grayt]$ of size $(3K,H,W)$
%\begin{align}
%    \begin{bmatrix} \u \\ \v \end{bmatrix} \hspace{-3pt} (t) \in \R^{3K \times H \times W}
%    \dot{\begin{bmatrix} \u \\ \v \end{bmatrix}} \hspace{-1pt} (t) = \begin{bmatrix} \mathrm{advection} \\ f_\theta \end{bmatrix} \in \R^{3K \times H \times W}
%\end{align}
%solved in one forward pass (See Appendix XX). We use finite differences to approximate the gradients $\nabla \u$ and $\nabla \v$.

\subsection{Modeling local and global effects}
PDEs link acceleration $\dv\grayxt$ solely to the current state and its gradient at the same location $\x$ and time $t$, ruling out long-range connections. However, long-range interactions naturally arise as information propagates over time across substantial distances. For example, Atlantic weather conditions influence future weather patterns in Europe and Africa, complicating the covariance relationships between these regions. Therefore, we propose a hybrid network to account for both local transport and global effects,
\begin{align} \label{eq:network}
    f_\theta\Big(\u\grayt, \nabla \u\grayt, \v\grayt, \psi\Big) = & \underbrace{f_\mathrm{conv}\Big( \u\grayt, \nabla \u\grayt, \v\grayt, \psi \Big)}_{\text{convolution network}} + \gamma\underbrace{f_\mathrm{att}\Big(\u\grayt, \nabla \u\grayt, \v\grayt, \psi\Big)}_{\text{attention network}}.
\end{align}

\paragraph{Local Convolutions} To capture local effects, we employ a \emph{local} convolution network, denoted as $f_\mathrm{conv}$. This network is parameterized using ResNets with 3x3 convolution layers, enabling it to aggregate weather information up to a distance of $L$ `pixels' away from the location $\x$, where $L$ corresponds to the network's depth. Additional parameterization details can be found in Appendix~\ref{sec:implement}.
\paragraph{Attention Convolutional Network} We include an attention convolutional network $f_\mathrm{att}$ which captures \emph{global} information by considering states across the entire Earth, enabling long-distance connections. This attention network is structured around KQV dot product, with Key, Query, and Value parameterized with CNNs. Further elaboration is provided in Appendix~\ref{subsec:att_conv} and $\gamma$ is a learnable hyper-parameter.
\begin{figure}[!t]
    \centering
    \includegraphics[width=\textwidth]{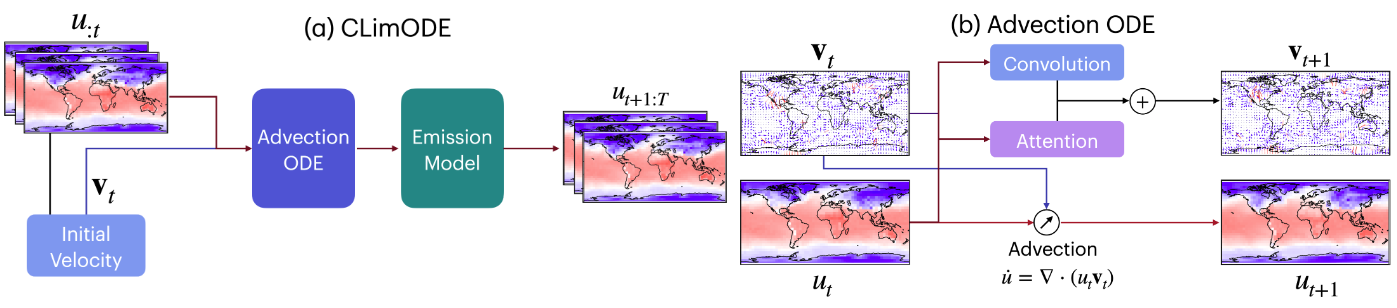}
    \caption{Whole prediction pipeline for ClimODE. }
    \label{fig:overview}
\end{figure}
\subsection{Spatiotemporal embedding} 
\paragraph{Day and Season} We encode daily and seasonal periodicity of time $t$ with trigonometric time embeddings \begin{align}
    \psi(t) = \bigg\{ \sin 2\pi t, \cos 2\pi t, \sin \frac{2\pi t}{365}, \cos \frac{2\pi t}{365} \bigg\}. %\quad \psi_{\mathrm{day}}(t) = \big( \sin 2\pi t, \cos 2\pi t \big), \quad \psi_{\mathrm{season}}(t) = \bigg( \sin \frac{2\pi t}{365}, \cos \frac{2\pi t}{365} \bigg).
\end{align}

\paragraph{Location} We encode latitude $h$ and longitude $w$ with trigonometric and spherical-position encodings
\begin{align}\label{eq:pos_enc}
    \psi(\x) = \big[ \{ \sin,\cos \} \times \{ h,w\}, \sin(h) \cos(w), \sin(h) \sin(w) \big].
\end{align}

\paragraph{Joint time-location embedding} We create a joint location-time embedding by combining position and time encodings ($\psi(t) \times \psi(\mathbf{x}$)), capturing the cyclical patterns of day and season across different locations on the map. Additionally, we incorporate constant spatial and time features, with $\psi(h)$ and $\psi(w)$ representing 2D latitude and longitude maps, and $\mathrm{lsm}$ and $\mathrm{oro}$ denoting static variables in the data,
\begin{align}
     \psi(\x,t) = \big[ \psi(t), \psi(\x), \psi(t) \times \psi(\x), \psi(c) \big], \qquad \psi(c) = \big[ \psi(h), \psi(w), \mathrm{lsm}, \mathrm{oro} \big]. 
\end{align}
%Combining all the above encodings constitutes the final set of spatiotemporal features. 
These spatiotemporal features are additional input channels to the neural networks (See Appendix~\ref{sec:data}).

%\begin{figure}[!t]
%    \centering
%    \includegraphics[width=\textwidth]{figs/fig_2_v3.png}
%    \caption{Temporal evolution of the system.}
%    \label{fig:change_system}
%\end{figure}

\subsection{Initial Velocity Inference} 
The neural transport model necessitates an initial velocity estimate, $\hat{\mathbf{v}}_k(\mathbf{x},t_0)$, to start the ODE solution \eqref{eq:solution}. In traditional dynamic systems, estimating velocity poses a challenging inverse problem, often requiring encoders in earlier neural ODEs \citep{chen2019neural,yildiz2019ode2vae,rubanova2019latent,de2019gru}. In contrast, the continuity Equation \eqref{eq:pde} establishes an identity, $\dot{u} + \nabla \cdot (u\mathbf{v}) = 0$, allowing us to solve directly for the missing velocity, $\mathbf{v}$, when observing the state $u$. We optimize the initial velocity for location $\x$, time $t$ and quantity $k$ with penalised least-squares
%to match the advection equation by penalized least-squares, where $\dot{u}$ is approximated by examining previous states $u(t < t_0)$ to obtain a numerical estimate of the change at $t_0$,
%We can easily approximate its time derivative, $\dot{u}$, by examining previous states $u(t < t_0)$ to obtain a numerical estimate of the change at $t_0$.
%The neural transport model requires an initial velocity estimate $\hat{\v}_k(\x,t_0)$ to start the ODE system \eqref{eq:solution}. In free-form dynamical systems, the velocity estimation is a challenging inverse problem, and earlier neural ODEs often resorted to encoders \citep{chen2019neural,yildiz2019ode2vae,rubanova2019latent,de2019gru}. In contrast, the continuity Equation \eqref{eq:pde} defines an identity $\dot{u} + \nabla \cdot (u\v) = 0$, from which we can directly solve the missing unknown $\v$ given that we observe the state $u$, and can easily approximate its time derivative $\dot{u}$ by looking at previous states $u(t < t_0)$ to obtain a numerical estimate of the change at $t_0$.
%Therefore, we utilize historical data points w.r.t to $\u_{t}$ such as $\u_{t-1},\u_{t-2}$ to obtain the best estimate of advection velocity for the whole spatial regime. 
%As a pre-processing step, we optimize the initial velocity for location $\x$, time $t$ and quantity $k$ to match the advection equation by penalized least-squares,
\begin{align}\label{eq:init_vel}
    \hat{\v}_k\grayt = \argmin_{\v_k\grayt} \:\: \bigg\{ \Big|\Big| \tilde{\dot{u}}_k\grayt + \v_k\grayt \cdot \tilde{\nabla} u_k\grayt + u_k\grayt \tilde{\nabla} \cdot \v_k\grayxt \Big|\Big|^2_2 + \alpha \big|\big| \v_k\grayt \big|\big|_\mathbf{K} \bigg\}, %\v\grayt^{\top} \mathbf{K} \v\grayt,
    %\operatorname{sum} \operatorname{tr}( d_\x \v\grayxt) 
\end{align}
where $\tilde{\nabla}$ is numerical spatial derivative, and $\tilde{\dot{u}}(t_0)$ is numerical approximation from the past states $u(t < t_0)$.
%are numerical derivatives over time or space. 
We include a Gaussian prior $\mathcal{N}(\operatorname{vec} \v_k | \0, \mathbf{K})$ with a Gaussian RBF kernel $\mathbf{K}_{ij} = \mathrm{rbf}(\x_i,\x_j)$ that results in spatially smooth initial velocities with smoothing coefficient $\alpha$. See Appendix~\ref{sec:intial_vel} for details.

%This optimisation problem is almost trivial since $\v$ is only two-dimensional. 

\subsection{System sources and uncertainty estimation}\label{sec:uncertain}

The model described so far has two limitations: (i) the system is deterministic and thus has no uncertainty, and (ii) the system is closed and does not allow value loss or gain (eg. during day-night cycle). We tackle both issues with an emission $g$ outputting a bias $\mu_k\grayxt$ and variance $\sigma_k^2\grayxt$ of $u_k\grayxt$ as a Gaussian,
\begin{align}
     u_k^\mathrm{obs}\grayxt \sim \N\Big( u_k\grayxt + \mu_k\grayxt, \sigma_k^2\grayxt \Big), \qquad \mu_k\grayxt,\sigma_k\grayxt = g_k\big( \u\grayxt, \psi \big).
\end{align}
The variances $\sigma_k^2$ represent the uncertainty of the climate estimate, while the mean $\mu_k$ represents value gain bias. For instance, the $\mu$ can model the fluctuations in temperature during the day-night cycle. This can be regarded as an emission model, accounting for the total aleatoric and epistemic variance. %The bias term models local value-preservation violations, such as night-time decrease and day-time increase in temperature. %In practise the error model addition improves the model performance significantly (See Section~\ref{sec:ablation}).

%We note that fully modelling the fluctuations would necessitate making the model significantly more complex. For instance, we could model atmosphere-earth transfers, or distribute the state into altitudes.  

\subsection{Loss}
We assume a full-earth dataset $\D = (\y_1, \ldots, \y_N)$ of a total of $N$ timepoints of observed frames $\y_i \in \R^{K \times H \times W}$ at times $t_i$. We assume the data is organized into a dense and regular spatial grid $(H,W)$, a common data modality. We minimize the negative log-likelihood of the observations $\y_i$,
\small
\begin{align}\label{eq:loss}
    \L(\theta ; \D) =  - \frac{1}{NKHW} \sum_{i=1}^N \bigg( \log \N\Big( \y_i | \u\grayti + \bmu\grayti, \operatorname{diag} \bs^2\grayti\Big) + \log \N_+\big( \bs\grayti | \0, \lambda_{\sigma}^2 I\big) \bigg),
    %\lambda ||\bs\grayti|| 
\end{align}
\normalsize
where we also add a Gaussian prior for the variances with a hypervariance $\lambda_{\sigma}$ to prevent variance explosion during training. We decay the $\lambda_{\sigma}^{-1}$ using cosine annealing during training to remove its effects and arrive at a maximum likelihood estimate. Further details are provided in Appendix~\ref{sec:train}.

\section{Experiments}
%\textcolor{red}{We consider research questions..}

\paragraph{Tasks} We assess ClimODE's forecasting capabilities by predicting the future state $\mathbf{u}_{t+\Delta t}$ based on the initial state $\mathbf{u}_t$ for lead times ranging from $\Delta t = 6$ to $36$ hours both global and regional weather prediction, and monthly average states for climate forecasting. Our evaluation encompasses global, regional and climate forecasting, as discussed in Sections \ref{sec:global}, \ref{sec:regional} and \ref{sec:climate}, focusing on key meteorological variables. %Additionally, we conduct a comprehensive ablation study to delve into the model's proficiency in capturing local and global patterns, shedding light on each component's role in shaping its overall performance in Section~\ref{sec:ablation}.
%We evaluate ClimODE on forecasting future state $\u_{t+\Delta t}$ given a starting state $\u_t$ for lead times of $\Delta t = (6, 12, \ldots, 36)$ hours. We evaluate global and regional forecasting in Section.~\ref{sec:global,sec:regional} over crucial meteorological variables. Additionally, we perform an in-depth ablation study to assess the behavior of model at modeling local vs global patterns as well as investigate the role played by each component of the model on its downstream performance.
%
%weather forecasting tasks:
%\begin{enumerate}
%    \item We evaluate the model to forecast the global weather at a future time $\u_{t+ \Delta t}$, given $\u_{t}$, across different lead-times $\Delta t$ in section~\ref{sec:global} and 
%    \item We also evaluate our method to forecast the regional-weather at a future time $\u_{t+ \Delta t}$, given $\u_{t}$, across different lead-times in section~\ref{sec:regional}.
%\end{enumerate} 
%Additionally, we perform ablation to assess each model component's effect on its downstream performance.

\begin{figure}[!t]
    \centering
    \includegraphics[width=\textwidth]{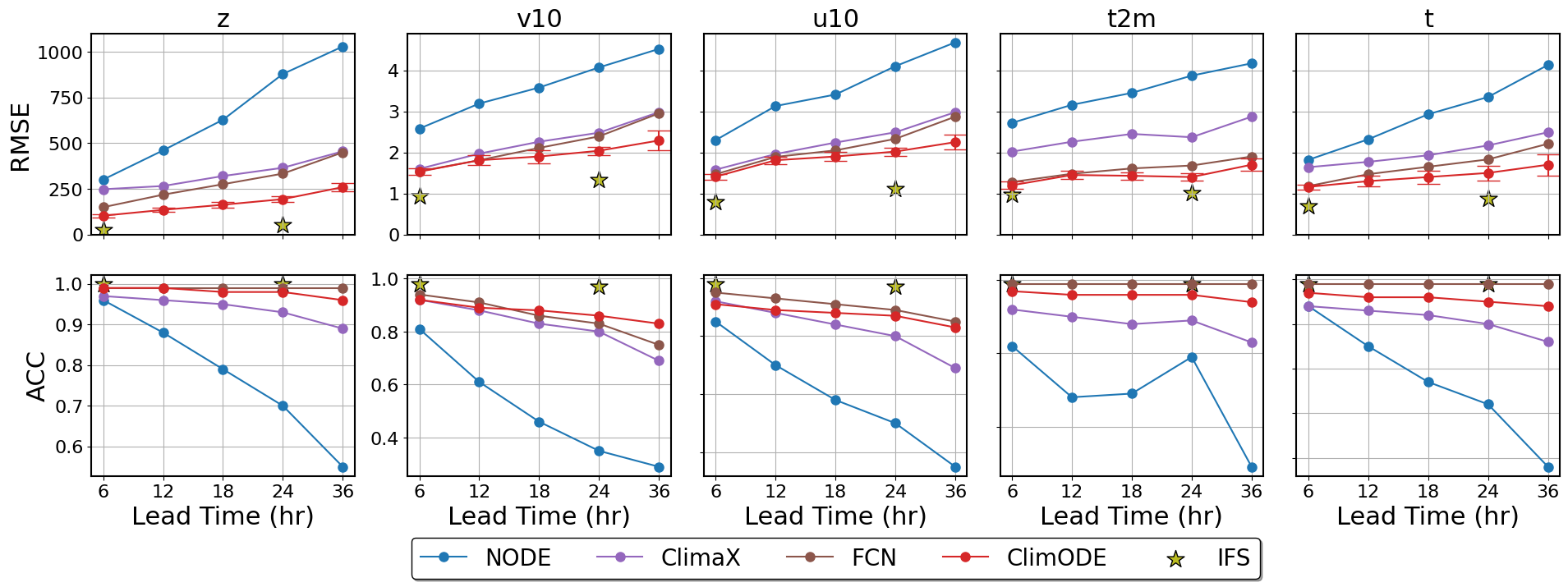}
    \caption{$\mathrm{RMSE} (\downarrow)$ and $\mathrm{ACC} (\uparrow)$ comparison with baselines. \textbf{ClimODE} outperforms competitive neural methods across different metrics and variables. For more details, see Table~\ref{tab:bench_table}.}
    \label{fig:global_forecast}
\end{figure}

\paragraph{Data.} We use the preprocessed $5.625^\circ$ resolution and 6 hour increment ERA5 dataset from WeatherBench \citep{rasp2020weatherbench} in all experiments. We consider $K=5$ quantities from the ERA5 dataset: ground temperature (\texttt{t2m}), atmospheric temperature (\texttt{t}), geopotential (\texttt{z}), and ground wind vector (\texttt{u10}, \texttt{v10}) and normalize the variables to $[0,1]$ via min-max scaling. Notably, both \texttt{z} and \texttt{t} hold standard importance as verification variables in medium-range Numerical Weather Prediction (NWP) models, while \texttt{t2m} and (\texttt{u10}, \texttt{v10}) directly pertain to human activities. We use ten years of training data (2006-15), the validation data is 2016 as validation, and two years 2017-18 as testing data. More details can be found in Appendix~\ref{sec:data}.
\paragraph{Metrics.} We assess benchmarks using latitude-weighted RMSE and Anomaly Correlation Coefficient (ACC) following the de-normalization of predictions.
\small
\begin{align}\label{eq:lat_rmse}
    \mathbf{\mathrm{RMSE}} = \frac{1}{N} \sum_t^{N} \sqrt{\frac{1}{H W} \sum_h^H \sum_w^W \alpha(h) (y_{thw} - u_{thw})^2},~&\mathbf{\mathrm{ACC}} = \frac{\sum_{t,h,w} \alpha(h) \tilde{y}_{thw} \tilde{u}_{thw}}{\sqrt{\sum_{t,h,w} \alpha(h) \tilde{y}^{2}_{thw}\sum_{t,h,w} \alpha(h) \tilde{u}^{2}_{thw}}}
\end{align}
\normalsize
where $\alpha(h) = \cos(h)/\frac{1}{H} \sum_{h'}^H \cos(h')$ is the latitude weight and $\tilde{y} = y - C$ and $\tilde{u} = u - C$ are averaged against empirical mean $C = \frac{1}{N} \sum_{t} y_{thw}$. More detail in Appendix~\ref{subsec:metric}.
%the latitude weighting factor defined as, where $\lambda_h$ is the latitude at position $h$, $C$ is the temporal mean of the ground truth data over the entire test set denoted as climatology, $C = \frac{1}{N}\sum_{i}y_{ihw}$,
%\begin{align}
%    w(h) = \frac{\cos(\lambda_{h})}{\frac{1}{H} \sum_{h^{\prime}}^{H} \cos(\lambda_{h^{\prime}})},~y^{\prime} = y^{\prime} - C,~ u^{\prime} = u^{\prime} - C
%\end{align}
%as benchmarks, defined as, where $N_{t}$ are the number of forecasts, $H$ and $W$ are latitude and longitude dimensions, $y_{ihw}$ is the truth  and $u_{ihw}$ is the predicted value for one observable,
\begin{table}[!t]
    \caption{$\mathrm{RMSE} (\downarrow)$ comparison with baselines for regional forecasting. \textbf{ClimODE} outperforms other competing methods in \texttt{t2m,t,z} and achieves competitive performance on \texttt{u10,v10} across all regions. }
    \label{tab:reg_bench}
    \centering
    \resizebox{0.9\textwidth}{!}{
    \begin{tabular}{lcccccc ccc ccc}
        \toprule
        & & \multicolumn{3}{c}{North-America} &  \multicolumn{3}{c}{South-America} &  \multicolumn{3}{c}{Australia}   \\
        \cmidrule(lr){3-5} \cmidrule(lr){6-8} \cmidrule(lr){9-11} 
        Value & Hours  &  NODE & ClimaX  &  ClimODE &  NODE & ClimaX  &  ClimODE &  NODE & ClimaX  &  ClimODE\\
        \midrule
        \multirow{4}{*}{z} & 6 & 232.8 & 273.4 &  \textbf{134.5} $\pm$ \textcolor{gray}{10.6}& 225.60 & 205.40 &  \textbf{107.7} $\pm$ \textcolor{gray}{20.2} & 251.4 & 190.2 &  \textbf{103.8} $\pm$ \textcolor{gray}{14.6}  \\
        & 12 & 469.2& 329.5 &  \textbf{225.0} $\pm$ \textcolor{gray}{17.3}& 365.6 & 220.15 &  \textbf{169.4} $\pm$ \textcolor{gray}{29.6} & 344.8 & 184.7 &  \textbf{170.7} $\pm$ \textcolor{gray}{21.0}\\
        & 18 &667.2  &543.0 &  \textbf{307.7} $\pm$ \textcolor{gray}{25.4}& 551.9  & 269.24 &  \textbf{237.8} $\pm$ \textcolor{gray}{32.2} & 539.9 & 222.2 &  \textbf{211.1} $\pm$ \textcolor{gray}{31.6}\\
        & 24 & 893.7  & 494.8   &  \textbf{390.1} $\pm$ \textcolor{gray}{32.3}& 660.3 & 301.81 &  \textbf{292.0} $\pm$ \textcolor{gray}{38.9}  & 632.7& 324.9 &  \textbf{308.2} $\pm$ \textcolor{gray}{30.6}\\
        \midrule
       \multirow{4}{*}{t} & 6 & 1.96 &1.62 &  \textbf{1.28} $\pm$ \textcolor{gray}{0.06}& 1.58  & 1.38 &  \textbf{0.97} $\pm$ \textcolor{gray}{0.13} &1.37 & 1.19  &  \textbf{1.05} $\pm$ \textcolor{gray}{0.12}\\
        & 12 & 3.34 & 1.86 &  \textbf{1.81} $\pm$ \textcolor{gray}{0.13}& 2.18 & 1.62 &  \textbf{1.25} $\pm$ \textcolor{gray}{0.18} & 2.18 & 1.30 &  \textbf{1.20} $\pm$ \textcolor{gray}{0.16}\\
        & 18 & 4.21 & 2.75 &  \textbf{2.03} $\pm$ \textcolor{gray}{0.16}& 2.74 & 1.79 &  \textbf{1.43} $\pm$ \textcolor{gray}{0.20} & 2.68& 1.39 &  \textbf{1.33} $\pm$ \textcolor{gray}{0.21}\\
        & 24 & 5.39 &2.27  &   \textbf{2.23} $\pm$ \textcolor{gray}{0.18}& 3.41  & 1.97 &  \textbf{1.65} $\pm$ \textcolor{gray}{0.26}  & 3.32 & 1.92 &  \textbf{1.63} $\pm$ \textcolor{gray}{0.24}\\
        \midrule
        \multirow{4}{*}{t2m} & 6 & 2.65 & 1.75 &  \textbf{1.61} $\pm$ \textcolor{gray}{0.2}& 2.12 & 1.85 &  \textbf{1.33} $\pm$ \textcolor{gray}{0.26}  & 1.88 & 1.57 &  \textbf{0.80} $\pm$ \textcolor{gray}{0.13}\\
        & 12 & 3.43 & \textbf{1.87} &  2.13 $\pm$ \textcolor{gray}{0.37}& 2.42 &2.08  &  \textbf{1.04} $\pm$ \textcolor{gray}{0.17} & 2.02 & 1.57 &  \textbf{1.10} $\pm$ \textcolor{gray}{0.22}\\
        & 18 & 3.53 & 2.27 &   \textbf{1.96} $\pm$ \textcolor{gray}{0.33}&2.60 & 2.15 &  \textbf{0.98} $\pm$ \textcolor{gray}{0.17} & 3.51 & 1.72 &  \textbf{1.23} $\pm$ \textcolor{gray}{0.24}\\
        & 24 & 3.39  & \textbf{1.93} &  2.15 $\pm$ \textcolor{gray}{0.20} & 2.56 & 2.23 &  \textbf{1.17} $\pm$ \textcolor{gray}{0.26} & 2.46  & 2.15 &  \textbf{1.25} $\pm$ \textcolor{gray}{0.25}\\
        \midrule
        \multirow{4}{*}{u10} & 6 &1.96  &1.74  & \textbf{1.54} $\pm$ \textcolor{gray}{0.19} & 1.94 & 1.27 &  \textbf{1.25} $\pm$ \textcolor{gray}{0.18} &1.91 & 1.40 &  \textbf{1.35} $\pm$ \textcolor{gray}{0.17}\\
        & 12 & 2.91 & 2.24 &  \textbf{2.01} $\pm$ \textcolor{gray}{0.20}& 2.74 & 1.57  &  \textbf{1.49} $\pm$ \textcolor{gray}{0.23} & 2.86& \textbf{1.77}  &  1.78 $\pm$ \textcolor{gray}{0.21}\\
        & 18 & 3.40 & 3.24 & \textbf{2.17} $\pm$ \textcolor{gray}{0.34}& 3.24 & 1.83 &  \textbf{1.81} $\pm$ \textcolor{gray}{0.29} & 3.44 & 2.03 &  \textbf{1.96} $\pm$ \textcolor{gray}{0.25}\\
        & 24 & 3.96 & 3.14 &  \textbf{2.34} $\pm$ \textcolor{gray}{0.32} & 3.77 & 2.04 &  \textbf{2.08} $\pm$ \textcolor{gray}{0.35} & 3.91 & 2.64 &  \textbf{2.33} $\pm$ \textcolor{gray}{0.33}\\
        \midrule
        \multirow{4}{*}{v10} & 6 & 2.16 &1.83 & \textbf{1.67} $\pm$ \textcolor{gray}{0.23} & 2.29 & 1.31 &  \textbf{1.30} $\pm$ \textcolor{gray}{0.21} &2.38 & 1.47 &  \textbf{1.44} $\pm$ \textcolor{gray}{0.20}\\
        & 12 & 3.20  & 2.43 &  \textbf{2.03} $\pm$ \textcolor{gray}{0.31}& 3.42  & \textbf{1.64} &  1.71 $\pm$ \textcolor{gray}{0.28} & 3.60 & \textbf{1.79} &  1.87 $\pm$ \textcolor{gray}{0.26}\\
        & 18 & 3.96 & 3.52 & \textbf{2.31} $\pm$ \textcolor{gray}{0.37}& 4.16 & \textbf{1.90} &  2.07 $\pm$ \textcolor{gray}{0.31} & 4.31& 2.33 &  \textbf{2.23} $\pm$ \textcolor{gray}{0.23}\\
        & 24 & 4.57 & 3.39 &  \textbf{2.50} $\pm$ \textcolor{gray}{0.41} & 4.76  & \textbf{2.14} &  2.43 $\pm$ \textcolor{gray}{0.34} & 4.88 & 2.58 &  \textbf{2.53} $\pm$ \textcolor{gray}{0.32}\\
        \bottomrule
    \end{tabular}}
\end{table}

\paragraph{Competing methods.} Our method is benchmarked against exclusively open-source counterparts. We compare primarily against \textbf{ClimaX}~\citep{nguyen2023climax}, a state-of-the-art Transformer method trained on same dataset, \textbf{FourCastNet (FCN)}~\citep{pathak2022fourcastnet}, a large-scale model based on adaptive fourier neural operators and against a \textbf{Neural ODE}. We were unable to compare with PanguWeather~\citep{bi2022panguweather}  and GraphCast~\citep{lam2022graphcast} due to unavailability of their code during the review period.
%Numerous deep learning methods, such as PanguWeather~\citep{bi2022panguweather} and GraphCast~\citep{lam2022graphcast}, lack publicly available open-source code, making it \emph{impossible for us to conduct comparisons with these approaches}. 
We ensure fairness by retraining all methods from scratch using identical data and variables without pre-training. 

\paragraph{Gold-standard benchmark.}
We also compare to the Integrated Forecasting System \textbf{IFS} \citep{ifs}, one of the most advanced global physics simulation model, often known as simply the `European model'. Despite its high computational demands, various machine learning techniques have shown superior performance over the IFS, as evidenced \citep{ben2024rise}, particularly when leveraging a multitude of variables and exploiting correlations among them, our study focuses solely on a limited subset of these variables, with IFS serving as the gold standard. More details can be found in Appendix~\ref{sec:train}.

\subsection{Global Weather Forecasting}
\label{sec:global}

\begin{figure}[!hbt]
    \centering
    \includegraphics[width=\textwidth]{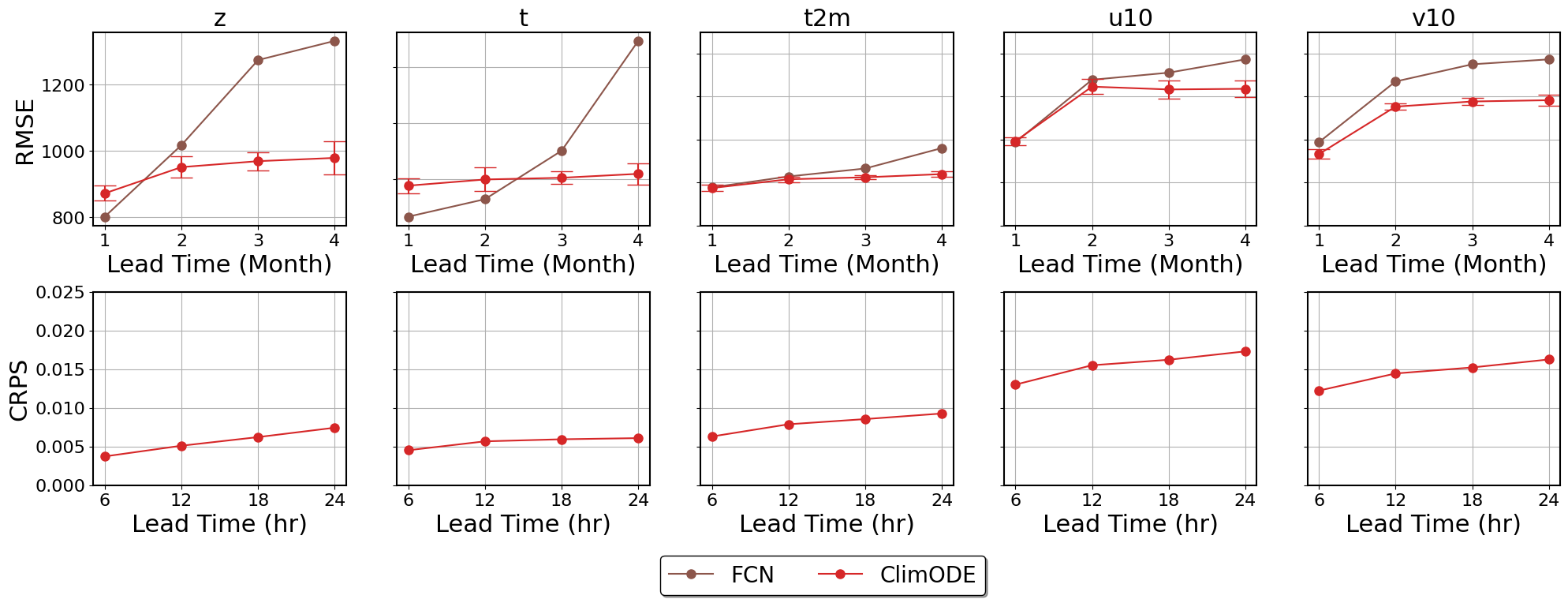}
    \caption{\textbf{CRPS and Monthly Forecasting}: $\mathrm{RMSE} (\downarrow)$ comparison with FourCastNet (FCN) for monthly forecasting and $\mathrm{CRPS}$ scores for ClimODE.  }
    \label{fig:crps_monthly}
\end{figure}

We assess ClimODE's performance in global forecasting, encompassing the prediction of crucial meteorological variables described above. Figure~\ref{fig:global_forecast} and Table~\ref{tab:bench_table} demonstrate ClimODE's superior performance across all metrics and variables over other neural baselines, while falling short against the gold-standard IFS, as expected. Fig. \ref{fig:crps_monthly} reports CRPS (Continuous Ranked Probability Score) over the predictions.These findings indicate the effectiveness of incorporating an underlying physical framework for weather modeling. %\color{red} Moreover, we also evaluated our model using CRPS (Continuous Ranked Probability Score) and Globaly monthly forecasting to forecast monthly averages of quantities, shown in Fig. \ref{fig:crps_monthly}. It can be seen our model can predict the weather quite well with predicted variance and bias outperform FCN in monthly forecasting. 
\color{black}

\subsection{Regional Weather Forecasting}
\label{sec:regional}
We assess ClimODE's performance in regional forecasting, constrained to the bounding boxes of %, which focuses on predicting future weather within specific regions based on current regional weather conditions. 
%Our evaluation encompasses 
North America, South America, and Australia, representing diverse Earth regions.  Table~\ref{tab:reg_bench} reveals noteworthy outcomes. ClimODE has superior predictive capabilities in forecasting ground temperature (\texttt{t2m}), atmospheric temperature (\texttt{t}), and geopotential (\texttt{z}). It also maintains competitive performance in modeling ground wind vectors (\texttt{u10} and \texttt{v10}) across these varied regions. This underscores ClimODE's proficiency in effectively modeling regional weather dynamics.

\subsection{Climate Forecasting: Monthly Average Forecasting}\label{sec:climate}
To demonstrate the versatility of our method, we assess its performance in climate forecasting. Climate forecasting entails predicting the average weather conditions over a defined period. In our evaluation, we focus on monthly forecasts, predicting the average values of key meteorological variables over one-month durations. We maintained consistency by utiliz the same ERA5 dataset and variables employed in previous experiments, and trained the model with same hyperparameters. Our comparative analysis with FourCastNet on latitude-weighted RMSE and ACC is illustrated in Figure~\ref{fig:crps_monthly}. Notably, ClimODE demonstrates significantly improved monthly predictions as compared to FourCastNet showing efficacy in climate forecasting.

\section{Ablation Studies}
\label{sec:ablation}

\begin{wrapfigure}[13]{r}{0.43\textwidth}
    \vspace{-25pt}
    %\hspace{-9pt}
    \includegraphics[scale=0.37, trim={0 0.1cm 0 0}, clip]{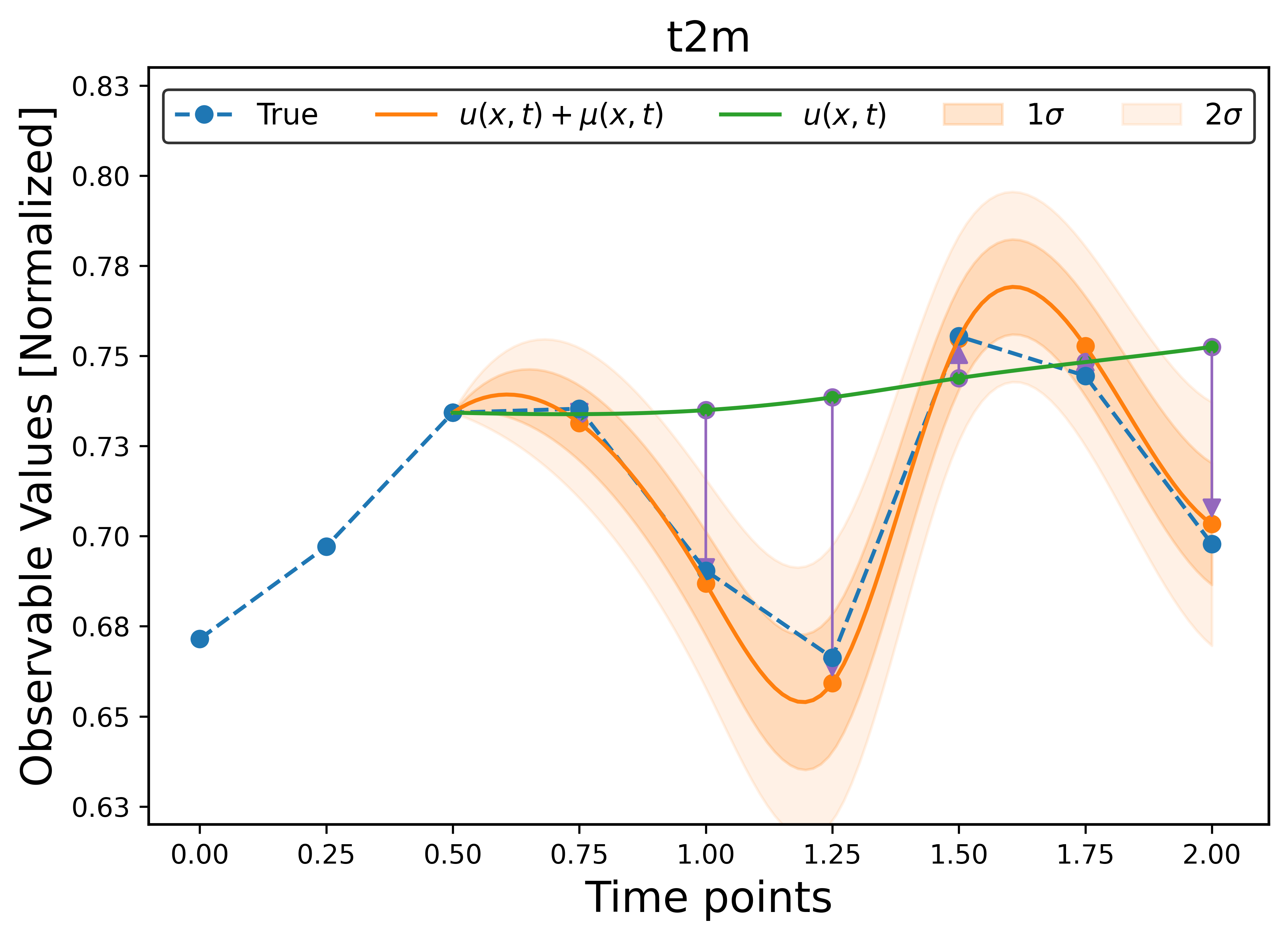}
    \caption{\textbf{Effect of bias}: \texttt{t2m} observed and predicted values showcasing the effect of bias.}
    \label{fig:bias}
\end{wrapfigure}

\begin{figure}[!hbt]
    \centering
    \includegraphics[width=\textwidth]{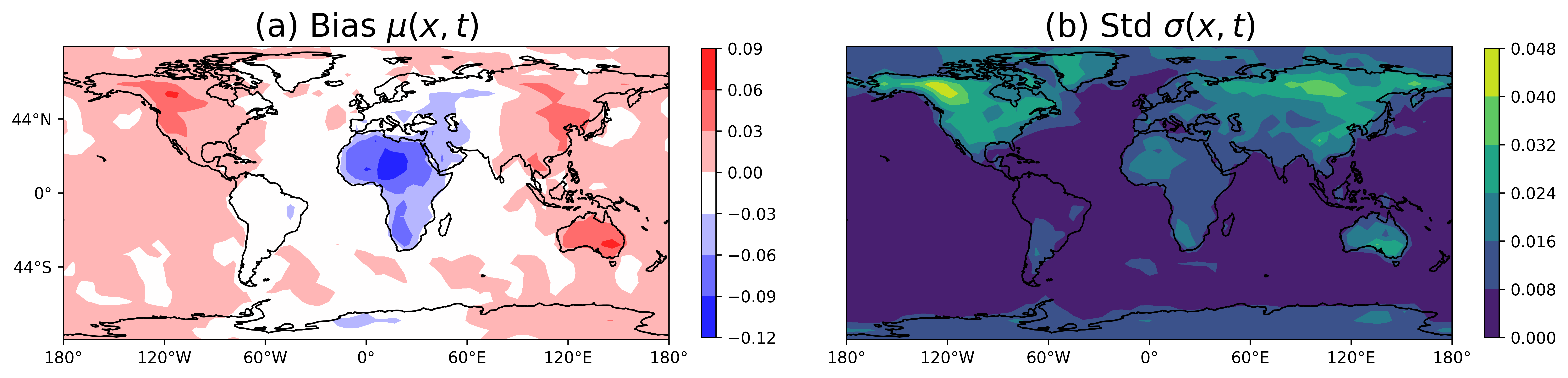}
    \caption{\textbf{Effect of emission model}: Global bias and standard deviation maps at 12:00 AM UTC. The bias explains day-night cycle \textbf{(a)}, while uncertainty is highest on land, and in north \textbf{(b)}.}
    \label{fig:day_night}
\end{figure}

\paragraph{Effect of emission model} %We assess the influence of the emission model, as outlined in Section~\ref{sec:uncertain}, both locally and globally. 
Figure~\ref{fig:bias} shows model predictions $u\grayxt$ of ground temperature (\texttt{t2m}) for a specific location while also including emission bias $\mu\grayxt$ and variance $\sigma^2\grayxt$. Remarkably, the model captures diurnal variations and effectively estimates variance. Figure \ref{fig:day_night} highlights bias and variance on a global scale. Positive bias is evident around the Pacific ocean, corresponding to daytime, while negative bias prevails around Europe and Africa, signifying nighttime. The uncertainties indicate confident ocean estimation, with northern regions being challenging. %This showcases the role of bias in learning day-night cycles locally and globally.

\begin{figure}[!hbt]
    \centering
    \includegraphics[width=\textwidth]{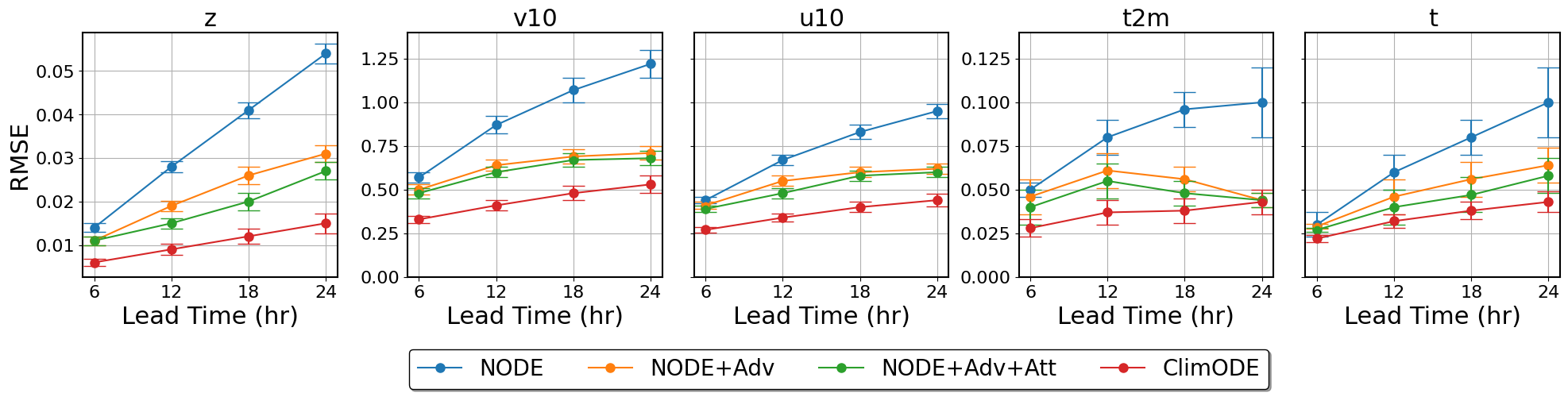}
    \caption{ \textbf{Effect of Individual Components}: The importance
of individual model components. An ablation showing how iteratively enhancing the vanilla neural ODE (\textcolor{blue}{blue}) with advection form (\textcolor{orange}{orange}), global attention (\textcolor{teal}{green}), and emission (\textcolor{red}{red}), improves performance of ClimODE. The advection component brings about the most accuracy improvements, while attention turns out to be least important.}
    \label{fig:ablation}
\end{figure}

\paragraph{Effect of individual components} We analyze the contributions of various model components to its performance. Figure \ref{fig:ablation} delineates the impact of components: \texttt{NODE} is a free-form second-order neural ODE, \texttt{Adv} corresponds to the advection ODE form, \texttt{Att} adds the attention in addition to convolutions, and \texttt{ClimODE} adds also the emission component. %We employ unweighted RMSE as our evaluation metric to compare these methods. 
All components bring performance improvements, with the advection and emission model having the largest, and attention the least effect. % Our findings reveal a discernible hierarchy of performance improvement by incorporating each component, underscoring the vital role played by each facet in enhancing the model's downstream performance. 
More details are in Appendix~\ref{sec:ablation_comp}.


\section{Conclusion and Future Work}

We present ClimODE, a novel climate and weather modeling approach implementing weather continuity. ClimODE precisely forecasts global and regional weather and also provides uncertainty quantification. While our methodology is grounded in scientific principles, it is essential to acknowledge its inherent limitations when applied to climate and weather predictions in the context of climate change. The historical record attests to the dynamic nature of Earth's climate, yet it remains uncertain whether ClimODE can reliably forecast weather patterns amidst the profound and unpredictable climate changes anticipated in the coming decades. Addressing this formidable challenge and also extending our method on newly curated global datasets ~\citep{rasp2023weatherbench} represents a compelling avenue for future research.

\section*{Acknowledgements} 
We thank the researchers at ECMWF for their open data sharing and maintenance of the ERA5 dataset, without which this work would not have been possible. We acknowledge CSC – IT Center for Science, Finland, for providing generous computational resources. This work has been supported by the Research Council of Finland under the {\em HEALED} project (grant 13342077). 

%YV thanks HTA (Helsinki Taitoluistelo Akatemi) for assistance in mental support and ice skating classes.  

%YV acknowledges support from the Research Council of Finland under the {\em HEALED} project (grant 13342077) and thanks Helsinki kaupunki Uusix for support and HTA for ice skating classes. 

%

%\section*{Reproducibility Statement}
%The Appendix~\ref{sec:data},\ref{sec:implement} provides extensive detail about the dataset used, the method's parametrization, and training details. Moreover, we also provide source code and model checkpoints in supplementary material. 

%\section*{Ethical and Societal Impact}

% Add in final version
%\paragraph{Acknowledgements}. funding

\bibliography{refs}
\bibliographystyle{iclr2024}

\newpage
\appendix

\section{Ethical Statement}
Deep learning surrogate models have the potential to revolutionize weather and climate modeling by providing efficient alternatives to computationally intensive simulations. These advancements hold promise for applications such as nowcasting, extreme event predictions, and enhanced climate projections, offering potential benefits like reduced carbon emissions and improved disaster preparedness while deepening our understanding of our planet.

\section{Data}
\label{sec:data}
We trained our model using the preprocessed version of ERA5 from WeatherBench~\citep{rasp2020weatherbench}. It is a standard benchmark data and evaluation framework for comparing data-driven weather forecasting models. WeatherBench regridded the original ERA5 at 0.25° to three lower resolutions: 5.625°, 2.8125°, and 1.40625°. We utilize the 5.625° resolution dataset for our method and all other competing methods. See \url{https://confluence.ecmwf.int/display/CKB/ERA5%3A+data+documentation}
for more details on the raw ERA5 data and Table~\ref{tab:data_tab} summarizes the variables used.

\begin{table}[!hbt]
\caption{ECMWF data variables in our dataset. \textit{Static} variables are time-independent, \textit{Single} represents surface-level variables, and \textit{Atmospheric} represents time-varying atmospheric properties at chosen altitudes.}
\label{tab:data_tab}
\begin{center}
\begin{tabular}{rllll}
\toprule
Type & Variable name & Abbrev. & ECMWF ID & Levels \\
\midrule
 Static & Land-sea mask & lsm & 172 & \\
Static & Orography & & & \\
Single & 2 metre temperature & t2m & 167 & \\
Single & 10 metre U wind component & u10 & 165 & \\
Single & 10 metre V wind component & v10 & 166 & \\
Atmospheric & Geopotential & z & 129 & $500$ \\
Atmospheric & Temperature & t & 130 & $850$ \\
\bottomrule
\end{tabular}
\end{center}
\end{table}

\subsection{Spherical geometry}
We model the data in a 2D latitude-longitude grid $\Omega$, but take the earth geometry into account by considering circular convolutions at the horizontal borders (international date line), and reflective convolutions at the vertical boundaries (north and south poles). We limit the data to latitudes $\pm 88^\circ$ to avoid the grid rows collapsing to the poles at $\pm 90^\circ$.

\section{Implementation Details}
\label{sec:implement}

\subsection{Model-Hyperparameters}

\begin{table}[!hbt]
\caption{Default hyperparameters for the emission model $g$}
\label{tab:hyper_error}
\begin{center}
\begin{tabular}{rlc}
\toprule
Hyperparameter & Meaning & Value \\
\midrule
Padding size & Padding size of each convolution layer & 1 \\
Padding type & Padding mode of each convolution layer & X: Circular, Y: Reflection \\
Kernel size & Kernel size of each convolution layer & 3 \\
Stride & Stride of each convolution layer & 1  \\
Residual blocks &  Number of residual blocks  & [3,2,2]  \\
Hidden dimension & Number of output channels of each residual block  & $[128,64,\text{out channels}]$\\
Dropout & Dropout rate & 0.1  \\
\bottomrule
\end{tabular}
\end{center}
\end{table}

\begin{table}[!hbt]
\caption{Default hyperparameters for the convolution network $f_{\mathrm{conv}}$}
\label{tab:hyper_acc}
\begin{center}
\begin{tabular}{rlc}
\toprule
Hyperparameter & Meaning & Value \\
\midrule
Padding size & Padding size of each convolution layer & 1 \\
Padding type & Padding mode of each convolution layer & X: Circular, Y: Reflection \\
Kernel size & Kernel size of each convolution layer & 3 \\
Stride & Stride of each convolution layer & 1  \\
Residual blocks &  Number of residual blocks  & [5,3,2]  \\
Hidden dimension & Number of output channels of each residual block  & $[128,64,\text{out channels}]$\\
Dropout & Dropout rate & 0.1  \\
\bottomrule
\end{tabular}
\end{center}
\end{table}

\subsection{Attention Convolutional Network}\label{subsec:att_conv}
We include an attention convolutional network $f_\mathrm{att}$ which captures \emph{global} information by considering states across the entire Earth, enabling the modeling of long-distance connections. This attention network is structured around Key-Query-Value dot product attention, with Key, Query, and Value maps parameterized as convolutional neural networks as,
\begin{itemize}
    \item \textbf{Key (K), Value (V)}: Key and Value maps are parameterized as 2-layer convolutional neural networks with stride=2 and $C_{K,V}$ as the latent embedding size. Based on the stride, this embeds every 4th pixel into a key, value latent vector of size $C_{K,V}$. We collect all embeddings into one tensor. 
    \item \textbf{Query (Q)}: Query map is parametrized as 2-layer convolutional neural networks with stride=1 and $C_{Q}$ as the latent embedding size. This incorporates somewhat local information and embeds into $C_{Q}$ latent vector. We collect all embeddings into one tensor.
\end{itemize}
We compute the attention maps via dot-product maps as,
\begin{align}
     \beta = \texttt{softmax}(QK^{\top})V
\end{align}
We consider a post-attention map for the attention coefficients as a 1-layer convolutional network with $1\times 1$ filter size to map the latent vectors into output channels.

\subsection{Metrics} \label{subsec:metric}
We assess benchmarks using latitude-weighted RMSE and Anomaly Correlation Coefficient (ACC) following the de-normalization of predictions.
\small
\begin{align}%\label{eq:lat_rmse}
    \mathbf{\mathrm{RMSE}} = \frac{1}{N} \sum_t^{N} \sqrt{\frac{1}{H W} \sum_h^H \sum_w^W \alpha(h) (y_{thw} - u_{thw})^2},~&\mathbf{\mathrm{ACC}} = \frac{\sum_{t,h,w} \alpha(h) \tilde{y}_{thw} \tilde{u}_{thw}}{\sqrt{\sum_{t,h,w} \alpha(h) \tilde{y}^{2}_{thw}\sum_{t,h,w} \alpha(h) \tilde{u}^{2}_{thw}}}
\end{align}
\normalsize
where $\alpha(h) = \cos(h)/\frac{1}{H} \sum_{h'}^H \cos(h')$ is the latitude weight and $\tilde{y} = y - C$ and $\tilde{u} = u - C$ are averaged against empirical mean $C = \frac{1}{N} \sum_{t} y_{thw}$. The anomaly correlation coefficient (ACC) gauges a model's ability to predict deviations from normal conditions. Higher ACC values signify better prediction accuracy, while lower values indicate poorer performance. It's a vital tool in meteorology and climate science for evaluating a model's skill in capturing unusual weather or climate events, aiding in forecasting system assessments. Latitude-weighted RMSE measures the accuracy of a model's predictions while considering the Earth's curvature. The weightage by latitude accounts for the changing area represented by grid cells at different latitudes, ensuring that errors in climate or spatial data are appropriately assessed. Lower latitude-weighted RMSE values indicate better model performance in capturing spatial or climate patterns.

%\subsection{Other implementation details}

\section{Training Details}\label{sec:train}

\subsection{Data normalization} 

We utilize 6-hourly forecasting data points from the ERA5 dataset and considered $K=5$ quantities from the ERA5 dataset: ground temperature (\texttt{t2m}), atmospheric temperature (\texttt{t}), geopotential (\texttt{z}), and ground wind vector (\texttt{u10}, \texttt{v10}) and normalize the variables to $[0,1]$ via min-max scaling. We use ten years of training data (2006--15), 2016 as validation data, and 2017--18 as testing data. There are 1460 data points per year and 2048 spatial points. 

\subsection{Data Batching}\label{sec:batching}

In our experiments, we utilize $K=5$ quantities (See Appendix~\ref{sec:data}) and spatial discretization of the earth to resolution $(H,W) = (32,64)$ resulting in a total of $3KWH = 30720$ scalar ODEs. This can seem daunting, but they all \emph{share the same differential function} $f_\theta$, that is, the time evolution at Tokyo and New York follows the same rules. The system can then be batched into a single image $\mathrm{stack}[\u\grayt;\v\grayt;\psi]$ of size $(3K+C,H,W)$, which is input to $f_{\theta}(\cdot):\R^{3K+C \times H \times W} \xrightarrow{} \R^{3K \times H \times W}$ and can be solved in one forward pass.
\begin{align}
    \begin{bmatrix} \u \\ \v \end{bmatrix} \hspace{-3pt} (t) \in \R^{(3K+C) \times H \times W}\quad 
    \dot{\begin{bmatrix} \u \\ \v \end{bmatrix}} \hspace{-1pt} (t) = \begin{bmatrix} \mathrm{advection} \\ f_\theta \end{bmatrix} \in \R^{3K \times H \times W}
\end{align}

We batch the data points wrt to years, giving the batch of shape $(N \times B \times 
(3K+C) \times H\times W)$, where $B$ is the batch size and $N$ here denotes the number of years. We used batch-size $B=8$ to train our model.

\subsection{Optimization}
\begin{figure}[!hbt]
    \centering
    \includegraphics[width=0.35\textwidth]{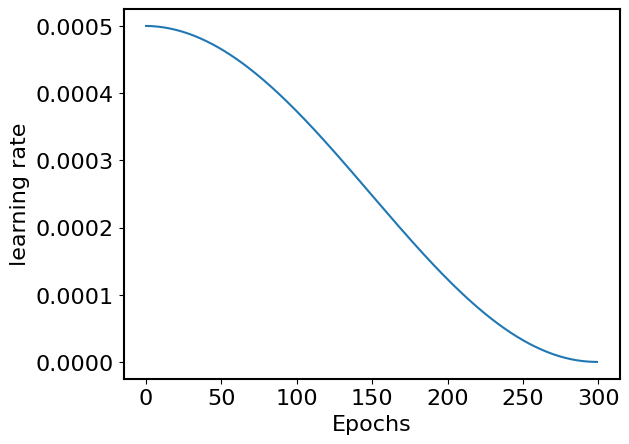}
    \includegraphics[width=0.35\textwidth]{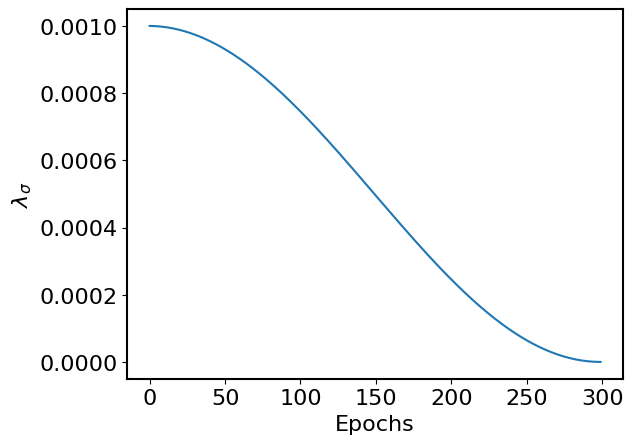}
    \caption{Learning rate and $\lambda_{\sigma}$ schedule wrt epochs.}
    \label{fig:lr_var}
\end{figure}

We used Cosine-Annealing-LR\footnote{\url{https://pytorch.org/docs/stable/generated/torch.optim.lr_scheduler.CosineAnnealingLR.html}} scheduler for the learning rate and also for the variance weight $\lambda_{\sigma}$ for L2 norm shown in Fig.~\ref{fig:lr_var} in the loss in Eq.~\ref{eq:loss}. We trained our model for 300 epochs, and the scheduler variation is shown below.

\subsection{Software and Hardware} 

The model is implemented in PyTorch~\citep{paszke2019pytorch} utilizing torchdiffeq~\citep{chen2019neural} to manage our data and model training. We use \texttt{euler} as our ODE-solver that solves the dynamical system forward with a time resolution of 1 hour. The whole model training and inference is conducted on a single 32GB NVIDIA V100 device.

\subsection{Initial velocity inference}\label{sec:intial_vel}
The neural transport model necessitates an initial velocity estimate, $\hat{\mathbf{v}}_k(\mathbf{x},t_0)$, to initiate the ODE system \eqref{eq:solution}. 
We estimate the missing velocity directly, $\mathbf{v}$, as a preprocessing step, for location $\x$, time $t$ and quantity $k$ to match the advection equation by penalized least-squares, where $\dot{u}$ is approximated by examining previous states $u(t < t_0)$ to obtain a numerical estimate of the change at $t_0$,
\begin{align}%\label{eq:init_vel}
    \hat{\v}_k\grayt = \argmin_{\v_k\grayt} \:\: \bigg\{ \Big|\Big| \tilde{\dot{u}}_k\grayt + \v_k\grayt \cdot \tilde{\nabla} u_k\grayt + u_k\grayt \tilde{\nabla} \cdot \v_k\grayxt \Big|\Big|^2_2 + \alpha \big|\big| \v_k\grayt \big|\big|_\mathbf{K} \bigg\}, %\v\grayt^{\top} \mathbf{K} \v\grayt,
    %\operatorname{sum} \operatorname{tr}( d_\x \v\grayxt) 
\end{align}
where $\{\tilde{\dot}, \tilde{\nabla}\}$ are numerical derivatives over time or space.
We compute $\tilde{\dot{u}}_k\grayt$ by utilizing torchcubicspline\footnote{\url{https://github.com/patrick-kidger/torchcubicspline}} to fit $\{\u_k\textcolor{gray}{(t-2)},\u_k\textcolor{gray}{(t-1)},\u_k\grayt\}$ to get a smooth derivative approximation. The spatial gradients $\tilde{\nabla}$ are calculated using \texttt{torch.gradient} function of PyTorch. We additionally place a Gaussian zero-mean prior $\mathcal{N}(\operatorname{vec} \v_k | \0, \mathbf{K})$ with a Gaussian RBF kernel $\mathbf{K}_{ij} = \mathrm{rbf}(\x_i,\x_j)$ that results in spatially smooth initial velocities with smoothing coefficient $\alpha$. The distance for the $\mathrm{rbf}(\x_i,\x_j)$ is computed as the euclidean norm between $\x_i$ and $\x_j$. This is optimized separately for each location $\x$ of the initial time $t_0$. We use Adam optimizer with a learning rate of 2 for 200 epochs. To get a balance between smoothing, local and global pattern we set the smoothing coefficient $\alpha=10^{-7}$.

%\section{Average Error Maps}

\section{Ablation Study Components}\label{sec:ablation_comp}
We conducted an extensive analysis to evaluate the individual contributions of each model component to its overall performance, as illustrated in Fig.~\ref{fig:ablation}. We delineate the impact of different components as,
\begin{itemize}
    \item \textbf{NODE}: A basic second-order neural differential equation as, here $f_{\mathrm{conv}}$ is parametrized by ResNet with the same set of parameters shown in Table~\ref{tab:hyper_acc},
    \begin{align}
        \dot{u}_k\grayxt &= \v_k\grayxt \\
        \dv_k\grayxt &= f_{\mathrm{conv}}\Big(\u\grayt, \nabla \u\grayt, \v\grayt, \psi\Big)
    \end{align}
    \item \textbf{NODE+Adv}: This combines the second-order neural differential equation with the advection component, where $f_{\mathrm{conv}}$ is parametrized by ResNet with the same set of parameters shown in Table~\ref{tab:hyper_acc},
    \begin{align}
        \dot{u}_k\grayxt &= - \v_k\grayxt \cdot \nabla u_k\grayxt - u_k\grayxt \nabla \cdot \v_k\grayxt \\
        \dv_k\grayxt &= f_{\mathrm{conv}}\Big(\u\grayt, \nabla \u\grayt, \v\grayt, \psi\Big)
    \end{align}
    \item \textbf{NODE+Adv+Att}: This the \texttt{NODE+Adv} with the attention convolutional network to model both local and global effects, where $f_{\mathrm{conv},\mathrm{att}}$ is parametrized by ResNet with the same set of parameters shown in Table~\ref{tab:hyper_acc} and Section~\ref{subsec:att_conv},
    \begin{align}
        \dot{u}_k\grayxt &= - \v_k\grayxt \cdot \nabla u_k\grayxt - u_k\grayxt \nabla \cdot \v_k\grayxt \\
        \dv_k\grayxt &= f_\mathrm{conv}\Big( \u\grayt, \nabla \u\grayt, \v\grayt, \psi \Big) + \gamma f_\mathrm{att}\Big(\u\grayt, \nabla \u\grayt, \v\grayt, \psi\Big)
    \end{align}
\end{itemize}

ClimODE encompasses all the previous components with the emission model component, including the bias and variance components. The $\texttt{NODE,NODE+Adv,NODE+Adv+Att}$ is trained by minimizing the MSE between predicted and truth observation as they output a point prediction and do not estimate uncertainty in the prediction. We employ unweighted RMSE as our evaluation metric to compare these methods. Our findings reveal a discernible hierarchy of performance improvement by incorporating each component, underscoring the vital role played by each facet in enhancing the model's downstream performance.

\begin{comment}

\subsection{Extension: latent modelling}

The current model only considers the $P$ observed quantities as explanatory variables. We need a way to account for unobserved or latent variables. A state-space model would consider a latent state and use encoders and decoders to move between ambient and latent space, and follows a general recipe
\begin{align}
    \u(t) &= \operatorname{dec} \circ \operatorname{pde} \circ \operatorname{enc}(\u_0).
\end{align}
\end{comment}

\section{Results Summary}

\begin{table}[!hbt]
    \caption{Latitude weighted $\mathrm{RMSE} (\downarrow)$ and $\mathrm{ACC} (\uparrow)$ comparison with baselines on global forecasting on ERA5 dataset.}
    \label{tab:bench_table}
    \centering
    \resizebox{\textwidth}{!}{
    \begin{tabular}{lccccccc ccccc}
        \toprule
        & & \multicolumn{5}{c}{$\mathrm{RMSE} (\downarrow)$} &  \multicolumn{5}{c}{$\mathrm{ACC}(\uparrow)$} \\
         \cmidrule(lr){3-7} \cmidrule(lr){8-12}
        Variable & Lead-Time (hours)  &  NODE & ClimaX & FCN  &  IFS & ClimODE &  NODE & ClimaX  & FCN & IFS & ClimODE \\
        \midrule
        %\multicolumn{2}{c}{$\#$ Parameters (M)} & $\sim 1.8$ & 107 &\textcolor{gray}{(N/A)} & $\sim 2.8$ \\ 
        %\midrule
        %& \atMod (2D-EGNN) & ($93.8 \pm 0.1$) & $93.8 \pm 0.1$ & 99.9 & 100   \\
        \multirow{4}{4em}{z} & 6 & 300.64 & 247.5 & 149.4 &26.9 & 102.9 $\pm$\textcolor{gray}{ 9.3} & 0.96 & 0.97& 0.99 &1.00& 0.99 \\
        & 12 & 460.23 & 265.3 & 217.8&\textcolor{gray}{(N/A)}  & 134.8 $\pm$ \textcolor{gray}{12.3}& 0.88& 0.96& 0.99 &\textcolor{gray}{(N/A)}& 0.99\\
        & 18 & 627.65 &319.8 & 275.0&\textcolor{gray}{(N/A)}  & 162.7 $\pm$ \textcolor{gray}{14.4}& 0.79& 0.95& 0.99 &\textcolor{gray}{(N/A)}& 0.98\\
        & 24 & 877.82 &364.9   & 333.0&51.0  & 193.4 $\pm$ \textcolor{gray}{16.3} & 0.70 & 0.93 & 0.99 & 1.00 & 0.98\\
        & 36 & 1028.20 &455.0   & 449.0&\textcolor{gray}{(N/A)}  & 259.6 $\pm$ \textcolor{gray}{22.3}& 0.55  & 0.89& 0.99 &\textcolor{gray}{(N/A)}& 0.96\\
        \midrule
       \multirow{4}{4em}{t} & 6 & 1.82 &1.64  & 1.18 & 0.69 & 1.16 $\pm$ \textcolor{gray}{0.06}& 0.94& 0.94& 0.99 &0.99&0.97\\
        & 12 & 2.32 & 1.77 & 1.47 &\textcolor{gray}{(N/A)}  & 1.32 $\pm$ \textcolor{gray}{0.13}& 0.85& 0.93& 0.99 &\textcolor{gray}{(N/A)}&0.96\\
        & 18 & 2.93 & 1.93 & 1.65 &\textcolor{gray}{(N/A)}  & 1.47 $\pm$ \textcolor{gray}{0.16}&0.77& 0.92& 0.99 &\textcolor{gray}{(N/A)}&0.96\\
        & 24 & 3.35&2.17  & 1.83 &0.87  & 1.55 $\pm$ \textcolor{gray}{0.18}& 0.72& 0.90& 0.99 &0.99& 0.95\\
        & 36 &  4.13 &2.49  & 2.21 &\textcolor{gray}{(N/A)}  & 1.75 $\pm$ \textcolor{gray}{0.26}&0.58 & 0.86& 0.99 &\textcolor{gray}{(N/A)}& 0.94\\
        \midrule
        \multirow{4}{4em}{t2m} & 6 & 2.72 &2.02  & 1.28& 0.97 & 1.21 $\pm$ \textcolor{gray}{0.09}&0.82& 0.92& 0.99 &0.99&0.97 \\
        & 12 & 3.16 & 2.26 & 1.48 &\textcolor{gray}{(N/A)}  &  1.45 $\pm$ \textcolor{gray}{0.10}&0.68&0.90& 0.99 &\textcolor{gray}{(N/A)}&0.96\\
        & 18 & 3.45 & 2.45 & 1.61 &\textcolor{gray}{(N/A)}  & 1.43 $\pm$ \textcolor{gray}{0.09}&0.69&0.88& 0.99 &\textcolor{gray}{(N/A)}&0.96\\
        & 24 & 3.86 & 2.37 & 1.68 &1.02  & 1.40 $\pm$ \textcolor{gray}{0.09}&0.79&0.89& 0.99 &0.99& 0.96\\
        & 36 & 4.17 &2.87  & 1.90 &\textcolor{gray}{(N/A)}  & 1.70 $\pm$ \textcolor{gray}{0.15}&0.49&0.83& 0.99 &\textcolor{gray}{(N/A)}& 0.94\\
        \midrule
        \multirow{4}{4em}{u10} & 6 & 2.3 &1.58 & 1.47 &0.80 & 1.41 $\pm$ \textcolor{gray}{0.07} &0.85& 0.92&0.95&0.98&0.91\\
        & 12 & 3.13 & 1.96 & 1.89 &\textcolor{gray}{(N/A)}  & 1.81 $\pm$ \textcolor{gray}{0.09}&0.70& 0.88 &0.93&\textcolor{gray}{(N/A)}&0.89\\
        & 18 & 3.41 & 2.24 & 2.05  &\textcolor{gray}{(N/A)}  & 1.97 $\pm$ \textcolor{gray}{0.11}&0.58& 0.84&0.91&\textcolor{gray}{(N/A)}&0.88\\
        & 24 & 4.1 & 2.49 & 2.33 &1.11  & 2.01 $\pm$ \textcolor{gray}{0.10}&0.50& 0.80&0.89&0.97&0.87 \\
        & 36 & 4.68 &2.98  & 2.87  &\textcolor{gray}{(N/A)}  & 2.25 $\pm$ \textcolor{gray}{0.18} &0.35&0.69& 0.85&\textcolor{gray}{(N/A)}&0.83\\
        \midrule
        \multirow{4}{4em}{v10} & 6 & 2.58 &1.60 & 1.54 &0.94 & 1.53 $\pm$ \textcolor{gray}{0.08}&0.81&0.92&0.94&0.98&0.92 \\
        & 12 & 3.19 & 1.97 & 1.81 &\textcolor{gray}{(N/A)}  & 1.81 $\pm$ \textcolor{gray}{0.12}&0.61&0.88&0.91&\textcolor{gray}{(N/A)}&0.89\\
        & 18 & 3.58 & 2.26 & 2.11 &\textcolor{gray}{(N/A)}  & 1.96 $\pm$ \textcolor{gray}{0.16}&0.46&0.83&0.86 &\textcolor{gray}{(N/A)}&0.88\\
        & 24 & 4.07 & 2.48 & 2.39&1.33 & 2.04 $\pm$ \textcolor{gray}{0.10}&0.35&0.80& 0.83&0.97& 0.86\\
        & 36 & 4.52 &2.98  & 2.95 &\textcolor{gray}{(N/A)}  & 2.29 $\pm$ \textcolor{gray}{0.24}&0.29&0.69& 0.75&\textcolor{gray}{(N/A)}& 0.83\\
        \bottomrule
    \end{tabular}}
\end{table}

\section{Longer Horizon Predictions}
Table~\ref{tab:bench_table_long} showcases the comparison of our method with ClimaX for for 72 hours (3 days) and 144 hours (6 days) lead time on latitude weighted RMSE  and ACC metrics. We observe that the temperature and potential (t,t2m,z) are relatively stable over longer forecasts, while the wind direction (u10,v10) becomes unreliable over a long time, which is an expected result. ClimaX is also remarkably stable over long predictions but has lower performance.
\begin{table}[!hbt]
    \caption{ \textbf{Longer lead time predictions}: Latitude weighted $\mathrm{RMSE} (\downarrow)$ and $\mathrm{ACC} (\uparrow)$ for longer lead times in global forecasting using the ERA5 dataset, in comparison with ClimaX.}
    \label{tab:bench_table_long}
    \centering
    \resizebox{0.6\textwidth}{!}{
    \begin{tabular}{lccc ccc}
        \toprule
        & & \multicolumn{2}{c}{$\mathrm{RMSE} (\downarrow)$} &  \multicolumn{2}{c}{$\mathrm{ACC}(\uparrow)$} \\
        \cmidrule(lr){3-4} \cmidrule(lr){5-6}
        Variable & Lead-Time (hours)  &  ClimaX  &  ClimODE &  ClimaX  &  ClimODE \\
        \midrule
        %\multicolumn{2}{c}{$\#$ Parameters (M)} & $\sim 1.8$ & 107 &\textcolor{gray}{(N/A)} & $\sim 2.8$ \\ 
        %\midrule
        %& \atMod (2D-EGNN) & ($93.8 \pm 0.1$) & $93.8 \pm 0.1$ & 99.9 & 100   \\
        \multirow{2}{4em}{z} & 72 &  687.0 & 478.7 $\pm$\textcolor{gray}{ 48.3} &  0.73& 0.88 $\pm$\textcolor{gray}{ 0.04} \\
        & 144 &  801.9 & 783.6 $\pm$\textcolor{gray}{ 37.3} &  0.58& 0.61  $\pm$\textcolor{gray}{ 0.13} \\
        \midrule
      \multirow{2}{4em}{t} & 72 &  3.17 &  2.58 $\pm$\textcolor{gray}{ 0.16} &  0.76& 0.85 $\pm$\textcolor{gray}{ 0.06} \\
        & 144 &  3.97 & 3.62 $\pm$\textcolor{gray}{ 0.21} &  0.69& 0.77  $\pm$\textcolor{gray}{ 0.16} \\
        \midrule
        \multirow{2}{4em}{t2m} & 72 &  2.87 &  2.75 $\pm$\textcolor{gray}{ 0.49} &  0.83&0.85 $\pm$\textcolor{gray}{ 0.14}  \\
        & 144 &  3.38 & 3.30 $\pm$\textcolor{gray}{ 0.23} &  0.83& 0.79 $\pm$\textcolor{gray}{ 0.25}\\
        \midrule
        \multirow{2}{4em}{u10} & 72 &  3.70 &  3.19$\pm$\textcolor{gray}{ 0.18} &  0.45& 0.66 $\pm$\textcolor{gray}{ 0.04} \\
        & 144 &  4.24 & 4.02 $\pm$\textcolor{gray}{ 0.12} &  0.30& 0.35 $\pm$\textcolor{gray}{ 0.08}\\
        \midrule
        \multirow{2}{4em}{v10} & 72 &  3.80 & 3.30 $\pm$\textcolor{gray}{ 0.22} &  0.39& 0.63 $\pm$\textcolor{gray}{ 0.05}  \\
        & 144 &  4.42 & 4.24$\pm$\textcolor{gray}{ 0.10} &  0.25& 0.32 $\pm$\textcolor{gray}{ 0.11}\\
        \bottomrule
    \end{tabular}}
\end{table}
We see that our method achieve better performance as compared to ClimaX for longer horizon predictions. 

\section{Validity of Mass Conservation}
\label{subsec:mass_consv}
To empirically study this, we analyzed how our current model retains the mass-conservation assumption and computed the integrals $I_{k,t} = \int u_k(\mathbf{x},t)d\mathbf{x}$ over time and quantities. We discovered that the value is constant over time up to $10^{-12}$.

\begin{figure}[!hbt]
    \centering
    \includegraphics[width=\textwidth]{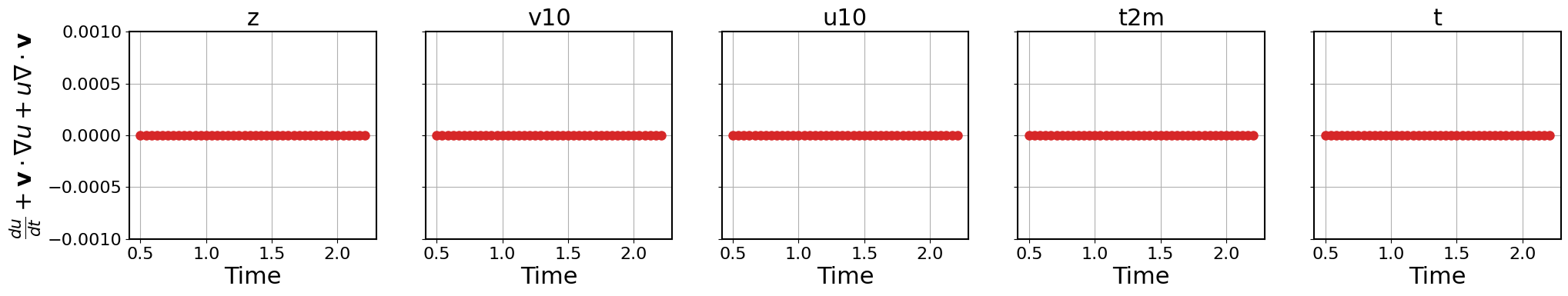}
    \caption{Validity of the mass conservation assumption of the ODE.}
    \label{fig:mass}
\end{figure}

\section{CRPS (Continuous Ranked Probability Score) and Climate Forecasting}
We further assessed our model using CRPS (Continuous Ranked Probability Score), as depicted in Figure \ref{fig:crps_monthly}. This analysis highlights our model's proficiency in capturing the underlying dynamics, evident in its accurate prediction of both mean and variance.

To showcase the effectiveness of our model in climate forecasting, we predicted average values over a one-month duration for key meteorological variables sourced from the ERA5 dataset: ground temperature (\texttt{t2m}), atmospheric temperature (\texttt{t}), geopotential (\texttt{z}), and ground wind vector (\texttt{u10}, \texttt{v10}). Employing identical data-preprocessing steps, normalization, and model hyperparameters as detailed in previous experiments, Figure \ref{fig:crps_monthly} illustrates the performance of ClimODE compared to FourCastNet in climate forecasting. Particularly noteworthy is our method's superior performance over FourCastNet at longer lead times, underscoring the multi-faceted efficacy of our approach.

\section{Correlation Plots}
To demonstrate the emerging couplings of quantities (ie. wind, temperature, pressure potential), we below plot the emission model  $\mathbf{u}^{\mathrm{pred}}(\mathbf{x},t) \in \mathbb{R}^5$ pairwise densities averaged over space $\mathbf{x}$ and time $t$. These effectively capture the correlations between quantities in the simulated weather states. These show that temperatures (t,t2m) and potential (z) are highly correlated and bimodal; the horizontal and vertical wind direction are independent (u10,v10); and there is little dependency between the two groups. These plots indicate that the emission model is highly aligned with data and does not indicate any immediate biases or skews. These results are averaged over space and time, and spatially local variations are still possible. The mean $\mu$ plots show that means match data well. The standard deviation $\sigma$ plots show some bimodality of predictions with either no or moderate uncertainty.
\begin{figure}[!hbt]
    \centering
    \includegraphics[width=\textwidth]{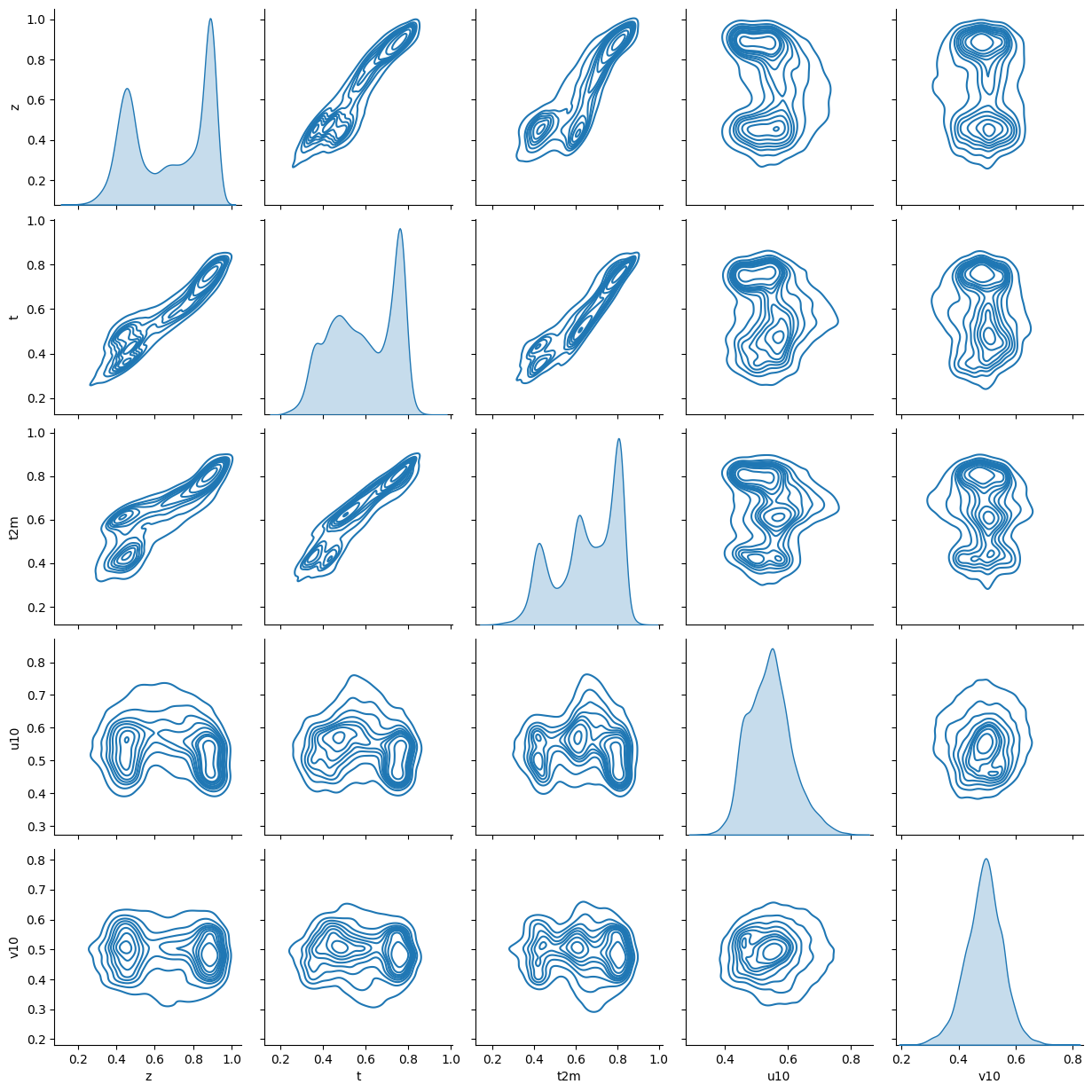}
    \caption{Pairwise correlation among the predicted variables by the model.}
    \label{fig:pair_cor}
\end{figure}

\begin{figure}[!hbt]
    \centering
    \includegraphics[width=0.25\textwidth]{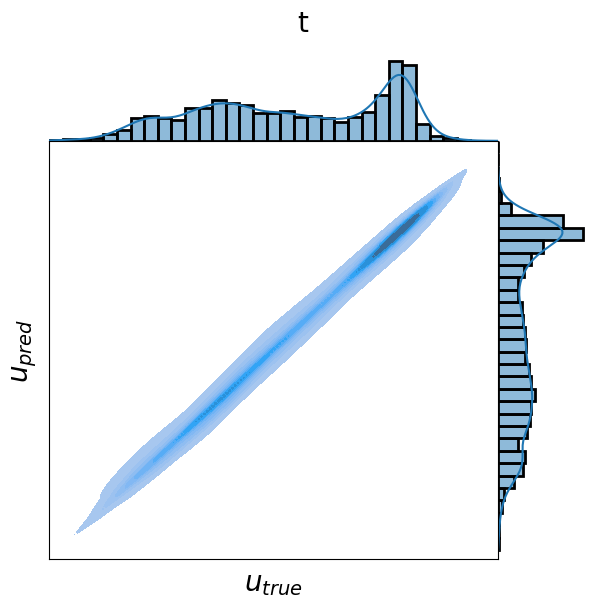}
    \includegraphics[width=0.25\textwidth]{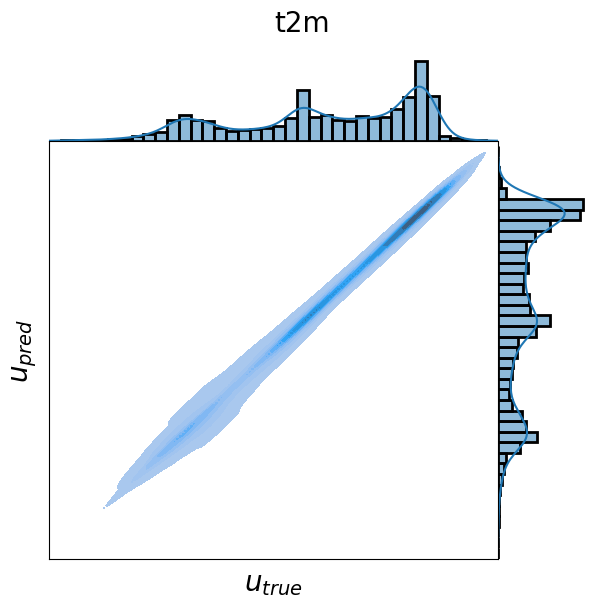}
    \includegraphics[width=0.25\textwidth]{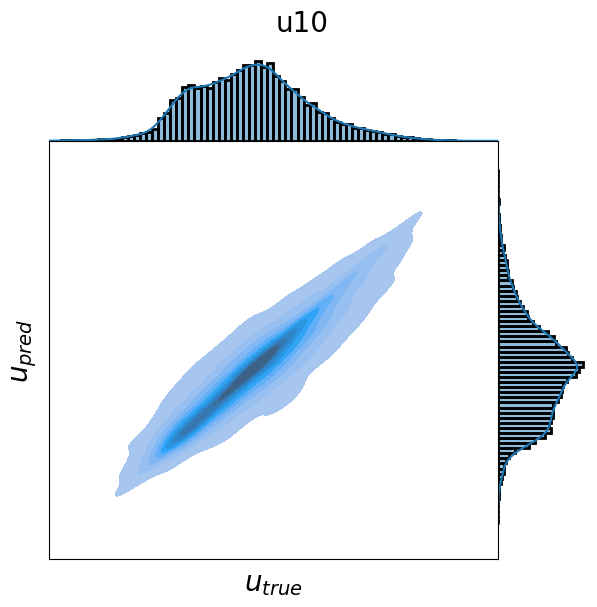}
    \includegraphics[width=0.25\textwidth]{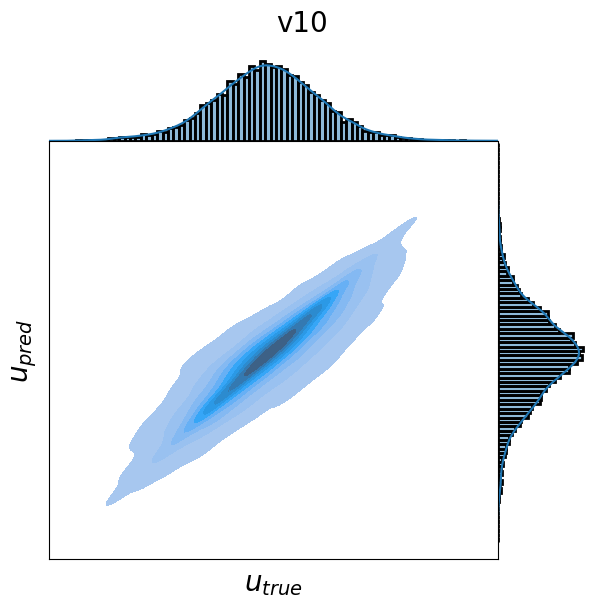}
    \includegraphics[width=0.25\textwidth]{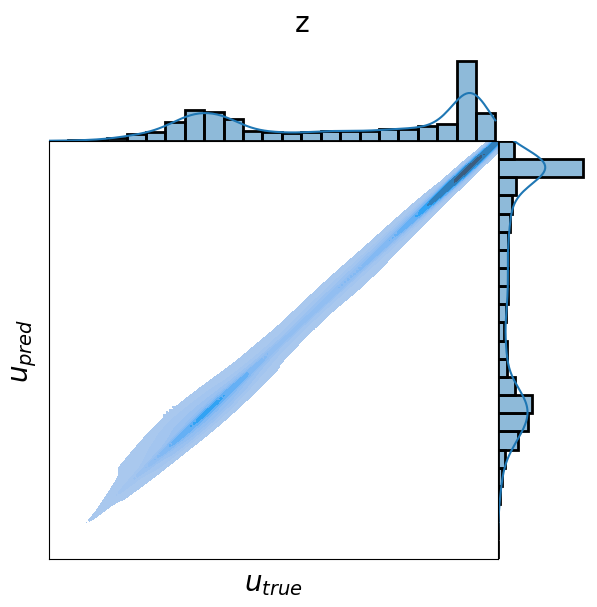}
    \caption{Correlation between $u_{pred}$ and $u_{true}$ for different observables, showing the efficacy of our model to predict the observables accurately.}
    \label{fig:corr_data}
\end{figure}

\begin{figure}[!hbt]
    \centering
    \includegraphics[width=0.25\textwidth]{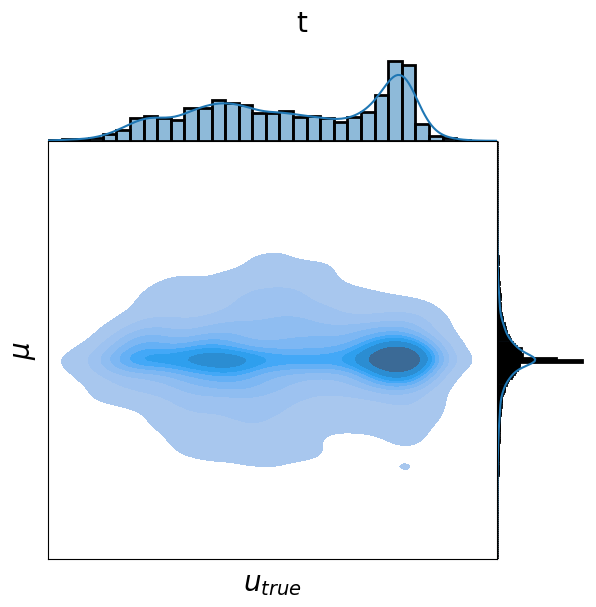}
    \includegraphics[width=0.25\textwidth]{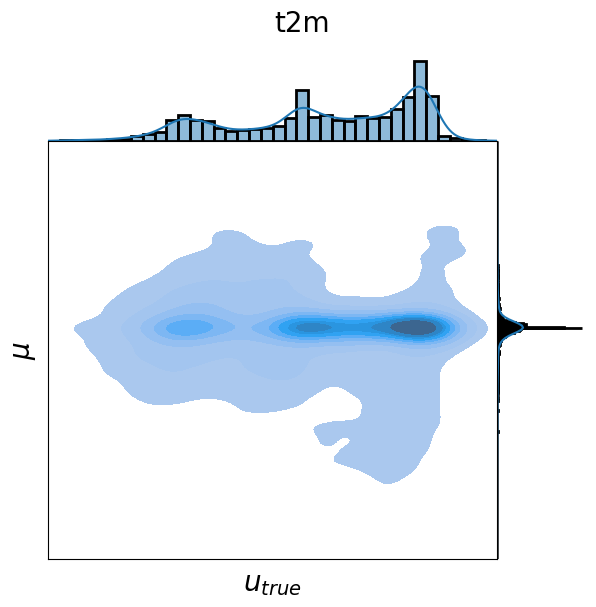}
    \includegraphics[width=0.25\textwidth]{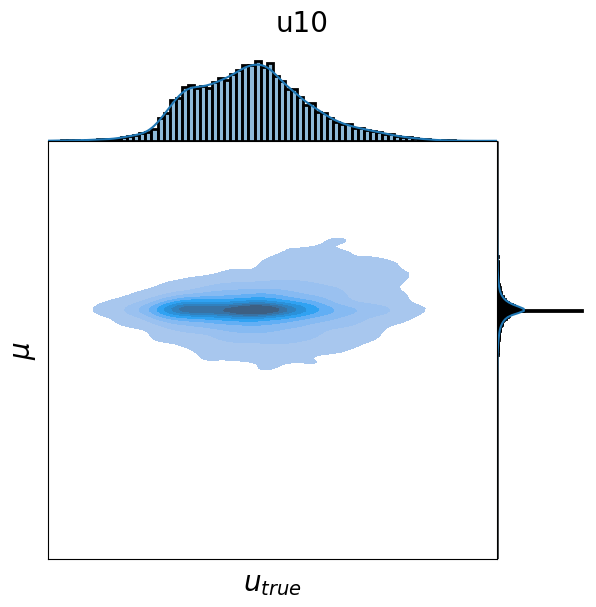}
    \includegraphics[width=0.25\textwidth]{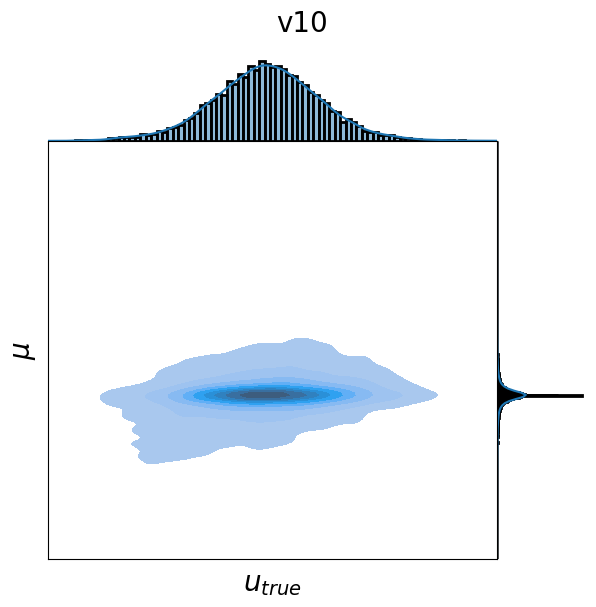}
    \includegraphics[width=0.25\textwidth]{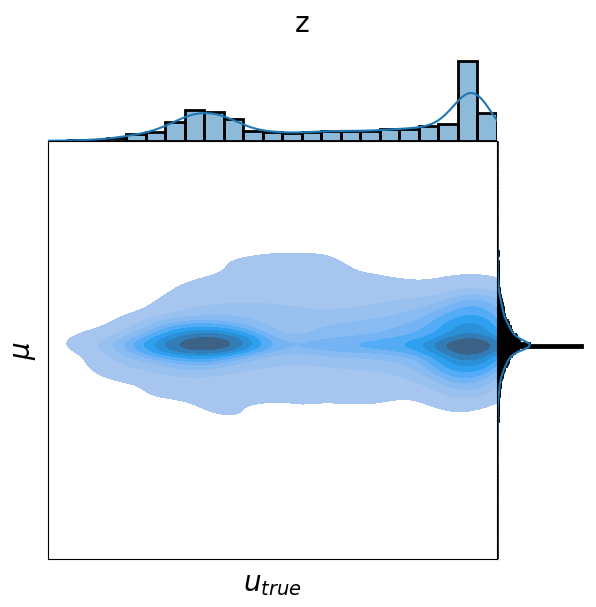}
\caption{Correlation between $\mu$ and $u_{true}$ for different observables.}
    \label{fig:corr_mu}
\end{figure}

\begin{figure}[!hbt]
    \centering
    \includegraphics[width=0.25\textwidth]{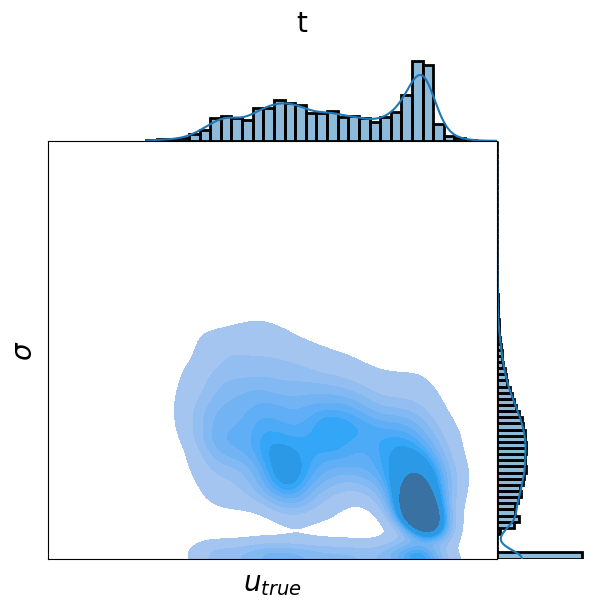}
    \includegraphics[width=0.25\textwidth]{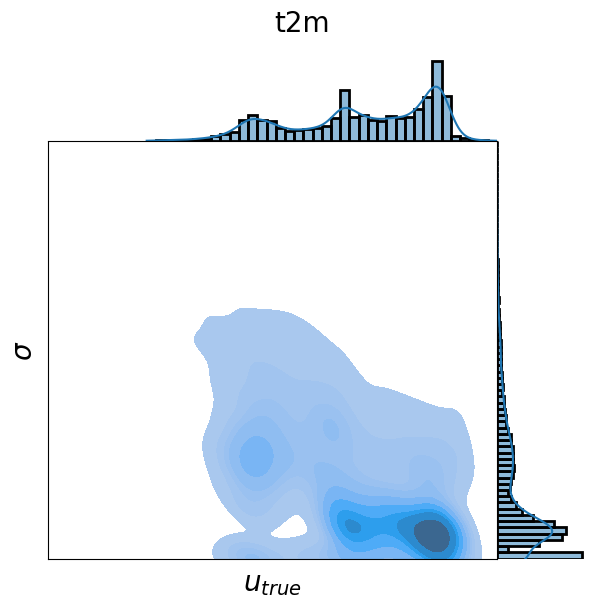}
    \includegraphics[width=0.25\textwidth]{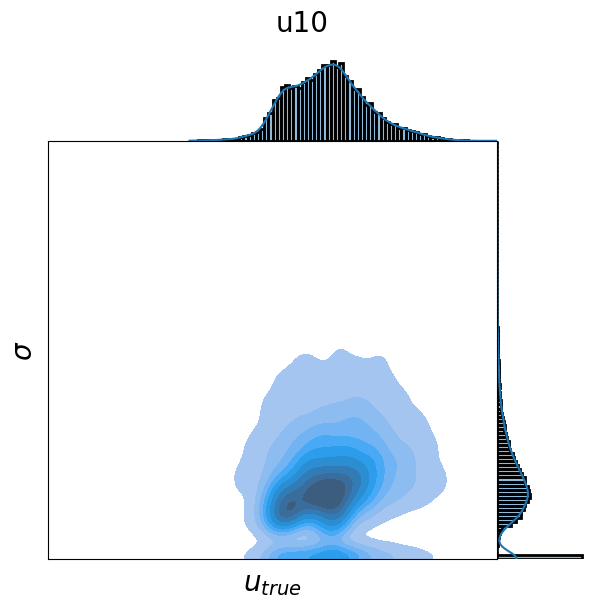}
    \includegraphics[width=0.25\textwidth]{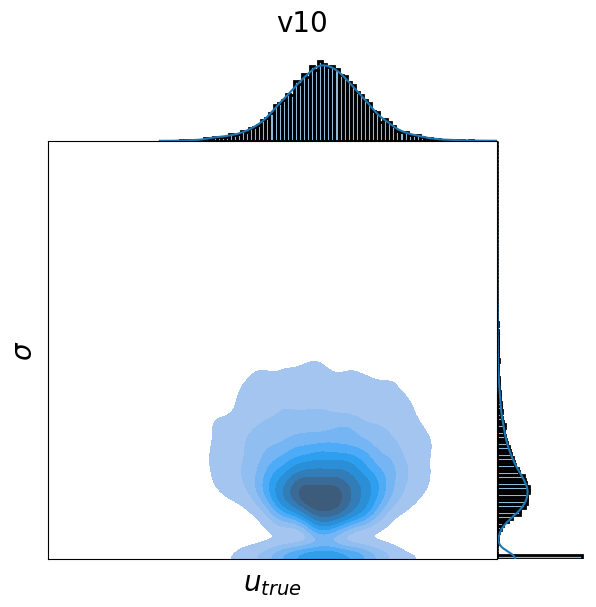}
    \includegraphics[width=0.25\textwidth]{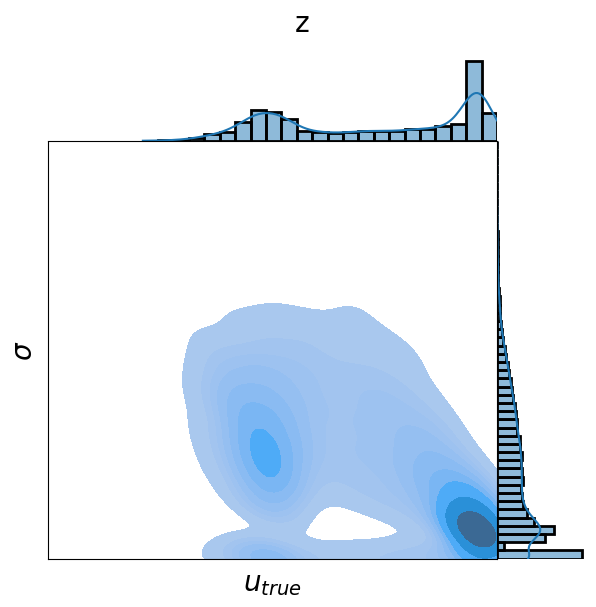}
\caption{Correlation between $\sigma$ and $u_{true}$ for different observables.}
    \label{fig:corr_sigma}
\end{figure}

\end{document}